\documentclass[lettersize,journal]{IEEEtran}
\usepackage{amsmath,amsfonts}
\usepackage{algorithmic}
\usepackage{algorithm}
\usepackage{array}
\usepackage[caption=false,font=normalsize,labelfont=sf,textfont=sf]{subfig}
\usepackage{textcomp}
\usepackage{stfloats}
\usepackage{url}
\usepackage{verbatim}
\usepackage{graphicx}
\usepackage{cite}
\usepackage{algorithm}  
\usepackage{algorithmic}  
\usepackage{microtype}
\usepackage{graphicx}
\usepackage{booktabs} % for professional tables

\usepackage{amssymb,mathrsfs,amsmath}
\usepackage{bbm}
\usepackage{multirow}
\usepackage{wrapfig}
\usepackage{amsthm,amsmath,amssymb}
\usepackage{mathrsfs}
\usepackage{color,xcolor}
\usepackage{amssymb}
\usepackage{bbding}
\usepackage{threeparttable}
\usepackage{balance}

\usepackage[colorlinks,
           linkcolor=red,
           filecolor=green, 
           urlcolor=blue,
	       citecolor=green,
           ]{hyperref}
\usepackage{amsmath}
\usepackage{amssymb}
\usepackage{mathtools}
\usepackage{amsthm}

\usepackage[capitalize,noabbrev]{cleveref}

\theoremstyle{plain}
\newtheorem{theorem}{Theorem}[section]

\newtheorem{lemma}[theorem]{Lemma}

\theoremstyle{definition}

\newtheorem{assumption}[theorem]{Assumption}
\theoremstyle{remark}

\newcommand{\eg}{{\em e.g.}}           % e.g.
\newcommand{\ie}{{\em i.e.}}           % i.e.
\definecolor{light-gray}{gray}{0.6}
\definecolor{light-gray-2}{gray}{0.9}

\begin{document}

\title{MutexMatch: Semi-supervised Learning with Mutex-based Consistency Regularization}


\author{Yue~Duan,
        Zhen~Zhao,
        Lei~Qi,
        Lei~Wang, \IEEEmembership{Senior Member,~IEEE,}
        Luping~Zhou, \IEEEmembership{Senior Member,~IEEE,}\\
        $\mbox{Yinghuan~Shi}$, and
        Yang~Gao, \IEEEmembership{Senior Member,~IEEE}% <-this % stops a space
\thanks{
% \IEEEcompsocthanksitem 
Y. Duan, Y. Shi and Y. Gao are with the National Key Laboratory for Novel Software Technology and the National Institute of Healthcare Data Science, Nanjing University, Nanjing 210023, China. E-mail: yueduan@smail.nju.edu.cn; 
\{syh, gaoy\}@nju.edu.cn. }
\thanks{
Z. Zhao and L. Zhou are with the School of Electrical and Information Engineering, The University of Sydney, Sydney, NSW 2006, Australia. E-mail:
\{zhen.zhao, luping.zhou\}@sydney.edu.au.}%\protect\\
\thanks{
L. Qi is with the School of Computer Science and Engineering and
the Key Lab of Computer Network and Information Integration (Ministry of Education), Southeast University, Nanjing 211189, China. E-mail: qilei@seu.edu.cn.}
\thanks{L. Wang is with the School of Computing and
Information Technology, University of Wollongong, Wollongong, NSW
2522, Australia. E-mail: leiw@uow.edu.au.}% <-this % stops an unwanted space
% \thanks{Manuscript received April 19, 2005; revised August 26, 2015.}
\thanks{This work was supported by NSFC Program (62222604, 62206052, 62192783), CAAI-Huawei MindSpore Project (CAAIXSJLJJ-2021-042A), China Postdoctoral Science Foundation Project (2021M690609), Jiangsu Natural Science Foundation Project (BK20210224), and CCF-Lenovo Bule Ocean Research Fund. (\textit{Corresponding author: Yinghuan Shi.})}
}

\markboth{ }%
{Shell \MakeLowercase{\textit{et al.}}: A Sample Article Using IEEEtran.cls for IEEE Journals}


\maketitle

\begin{abstract}
The core issue in semi-supervised learning (SSL) lies in how to effectively leverage unlabeled data, whereas most existing methods tend to put a great emphasis on the utilization of high-confidence samples yet seldom fully explore the usage of low-confidence samples.
% Early SSL methods mostly require low-confidence samples to optimize the same loss function as high-confidence samples, but this setting might largely challenge low-confidence samples especially at the early training stage. 
In this paper, we aim to utilize low-confidence samples in a novel way with our proposed mutex-based consistency regularization, namely \textit{MutexMatch}. Specifically, the high-confidence samples are required to exactly predict ``what it is" by conventional True-Positive Classifier, while low-confidence samples are employed to achieve a simpler goal --- to predict with ease ``what it is not'' by True-Negative Classifier. In this sense, we not only mitigate the pseudo-labeling errors but also make full use of the low-confidence unlabeled data by consistency of dissimilarity degree. MutexMatch achieves superior performance on multiple benchmark datasets, 
\ie, CIFAR-10, CIFAR-100, SVHN, STL-10,  mini-ImageNet and Tiny-ImageNet. 
More importantly, our method further shows superiority when the amount of labeled data is scarce, \eg, 92.23\% accuracy with only 20 labeled data on CIFAR-10. Our code and model weights have been released at \url{https://github.com/NJUyued/MutexMatch4SSL}.
\end{abstract}

\begin{IEEEkeywords}
Semi-supervised classification, mutex-based consistency regularization.
\end{IEEEkeywords}

\section{Introduction}
\label{sec:intro}
\IEEEPARstart{T}{o} avoid time-consuming and laborious labeling tasks, semi-supervised learning (SSL)~\cite{zhu2009introduction,chapelle2009semi,zhu2017semi,van2020survey}
% a longstanding and important field in machine learning field, 
is adopted to leverage a large quantity of unlabeled data along with a small quantity of labeled data during training. 
Recent semi-supervised learning (SSL) models could be categorized into two types: \textit{consistency regularization} based methods and \textit{entropy minimization} based methods, while the utilization of unlabeled data is crucial in both. In specific, consistency regularization based methods, \eg, ~\cite{laine2016temporal,tarvainen2017mean}, tend to utilize all unlabeled data together with the supervision from labeled data, which is at the risk of strong confirmation bias \cite{yu2018learning}. Although recent holistic methods, such as~\cite{sohn2020fixmatch,xu2021dash}, realize consistency regularization by integrating entropy minimization into the training process via pseudo-labeling \cite{lee2013pseudo}, they share a non-negligible limitation: they use only certain unlabeled data (\ie, with high confidence) to participate in the training while
% they all require a predefined confidence threshold to control whether certain unlabeled data should participate in training or not, 
preventing uncertain unlabeled data (\ie, with low confidence) from being fully exploited.
This waste of unlabeled samples may greatly deteriorate the outcome of final performance. In comparison, recent entropy minimization based method~\cite{rizve2021in} employs pseudo-labeling to incorporate partial samples selected by a low-confidence threshold into training process, while 
% in addition to possible error accumulation,
certain low-confidence samples are still neglected. 
% causes these conventional models difficult to learn the potential pattern from all unlabeled data, which 
% might deteriorate final performance.
The reason for discarding low-confidence samples in previous methods lies in these samples could be easily mispredicted. Thus,  the core problem is to tradeoff informatively learning low-confidence samples and preventing possible error accumulation.
\begin{figure}
   \centering
   \includegraphics[width=0.5\textwidth]{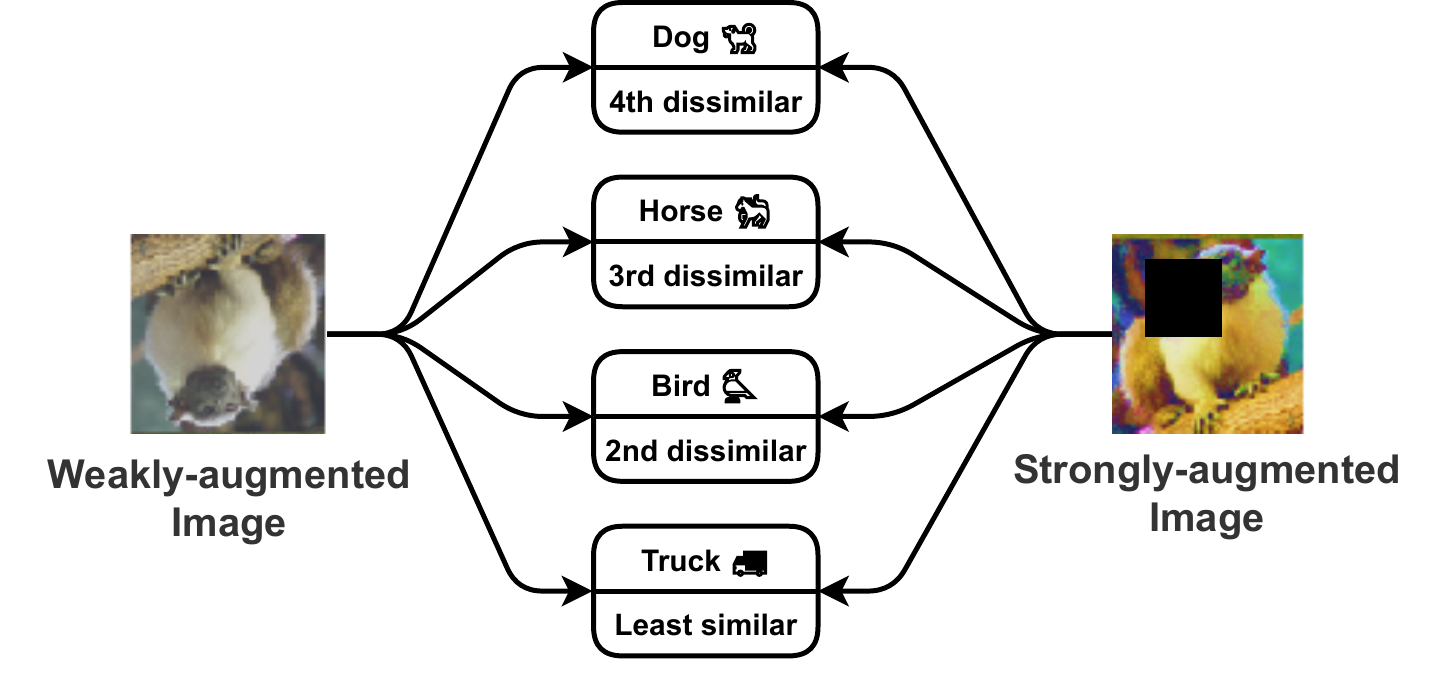}
%   \vspace{-1em}
   \caption{Given a monkey image, it is neither a dog, nor a bird, nor a horse, nor a truck. For the weakly-augmented version of this image, the model believes that it resembles a truck the least, then a horse, a bird, and a dog. For the strongly-augmented version of this image, 
   the order of dissimilarity degree could share some overlap.}
   \label{ex_fig}  
%   \vspace{-1em}
\end{figure}

Given these limitations, we try to answer---if we could treat these low-confidence unlabeled samples in an appropriate way? When imaging a low-confidence sample with its actual label of a ``horse", in some cases, it might be difficult for a trained model to make an accurate prediction, \ie, \textit{``it is a horse"}. On the contrary, it should be much easier for the model to ``\textit{guess what it is not}, \eg, \textit{it is not a cat}'' or ``\textit{it is not an elephant}''.
This drives us to consider a more straightforward yet feasible direction: for low-confidence samples, can we design a paradigm to learn from ``what it is not'' so as to benefit the learning of ``what it is".
In specific, \textbf{\textit{for different augmented versions of an image with the same semantic label, though their accurate predictions might be totally different, their predictions of the degree of dissimilarity could share some overlap}}. As illustrated in \cref{ex_fig} where there are many alternatives to answer what it is not, we believe the consistency of this dissimilarity degree can benefit the model to learn an informative representation especially for low-confidence samples. Thus, 
% to ensure a more informative feature learning from low-confidence samples, 
we  impose consistent prediction on two different augmented versions of the same low-confidence sample, both towards its complementary label, which is analogous to the consistency loss imposed on high-confidence samples commonly employed in consistency-based SSL. 
% That is to say, for a low-confidence image, it is unnecessary to get a certain class, and its complementary pseudo-label is easier to obtain. 
% Therefore, we could learn less error information when using unlabeled data with low-confidence.
In this sense, when using low-confidence unlabeled data with more easily obtained complementary labels, less error information could be expected to benefit the learning for ``what it is".

\begin{figure*}[t]  
\setlength{\abovecaptionskip}{-1em}
   \begin{center}
   %\framebox[4.0in]{$\;$}
   \resizebox{0.98\textwidth}{!}{          
      \includegraphics{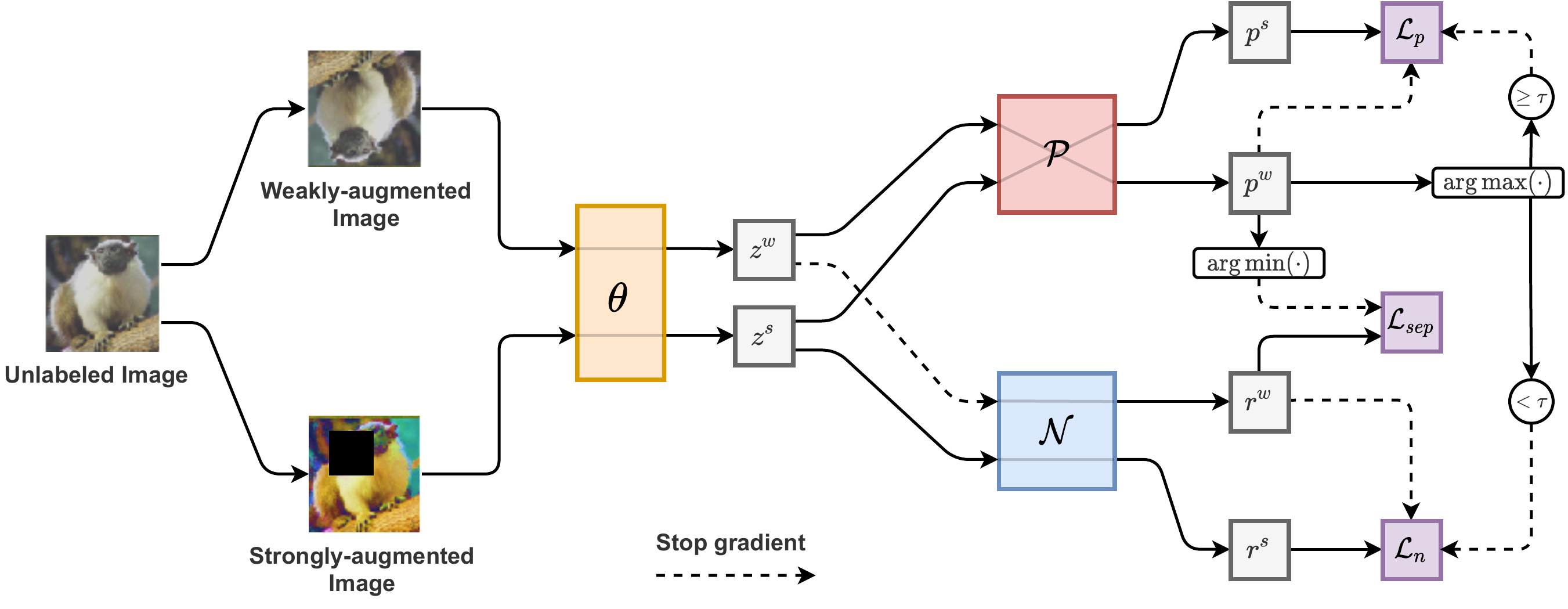}
   }  
   \end{center}
    %   \vspace{-0.5cm}
   \caption{The diagram of MutexMatch. Given a batch of unlabeled samples, 
   TPC $\mathcal{P}$ first uses their weakly-augmented variants to generate pseudo-labels. Then,   the classes with the lowest confidence are adopted as the complementary labels to train TNC $\mathcal{N}$ separately. 
   Meanwhile, TPC and TNC are used for mutex-based consistency regularization in the high and low-confidence portions of TPC's predictions respectively. $z$ denotes output features while $p$ and $r$ denote predictions of TPC and TNC. Superscripts $w$ and $s$ represent corresponding outputs for the weakly-augmented variants and strongly-augmented variants, respectively.}
   \label{fig1}
%   \vskip -0.04in
\end{figure*}

To achieve the aforementioned goal, we propose a novel framework, \textbf{MutexMatch}, to leverage all unlabeled data, especially low-confidence unlabeled data. We utilize \textit{True-Positive Classifier} \textbf{(TPC)}: to predict ``what it is'', and \textit{True-Negative Classifier} \textbf{(TNC)}: to predict ``what it is not'', which is designed to learn the feature representation of unlabeled data from a mutex perspective. 
Like conventional way in previous works~\cite{sohn2020fixmatch}, the weakly-augmented unlabeled samples are also used to generate pseudo-labels, while strong augmentation is achieved by using RandAugument~\cite{cubuk2020randaugment}. 
Specifically, we set a threshold to control the high-confidence portion and low-confidence portion of pseudo-labels. In high-confidence portion, we enforce the consistency regularization on the output of TPC. 
Compared with the existing methods, an improvement of MutexMatch is that it allows \textit{low-confidence} unlabeled samples to participate in training in a manner of \textit{complementary label} (\ie, a class of ``not belonging'') outputted from TNC, which helps model learn better data representation to further boost TPC. We find that the class with the smallest probability predicted by TPC already serves as a simple and reliable estimation of the complementary label to train TNC. Whereas, in low-confidence portion, we enforce consistency regularization on TNC with soft labels whose intensity is controlled by
a top-$k$ algorithm, \ie, we encourage the model to maintain a consistent degree of dissimilarity in the top-$k$ dissimilar classes where a larger $k$ represents a stricter consistency.  
By the above design, our approach is able to provide additional information from \textit{low-confidence} samples to the model in a \textit{relaxed} manner.
% This aforementioned mutex-based consistency could largely boost the conventional SSL learning by incorporating a previous ignored viewpoint---the consistency of dissimilar on low-confidence samples.
Finally, we theoretically analyze an objective function involving mutex-based consistency regularization. The diagram of proposed MutexMatch is shown in \cref{fig1}.

In this work, the key contributions include threefold: 
\begin{enumerate}
    \item[(1)] We use two classifiers (\ie, TPC and TNC) to construct \textit{MutexMatch}, a novel framework using pseudo-label and complementary label to learn an informative representation of unlabeled samples.  
    \item[(2)] We propose \textit{mutex-based consistency regularization} for SSL, which aims to make full use of unlabeled data, especially low-confidence unlabeled samples, in an effective way. We also theoretically show our error bound is lower than that of conventional consistency-based model without TNC.
    \item[(3)] We obtain superior results than recently-proposed SSL algorithms, \eg, on the most commonly studied SSL benchmark  CIFAR-10, MutexMatch using only 20 labels can achieve an accuracy of 92.23$\pm$3.23\%.
\end{enumerate}

In the rest of this paper, \cref{rework} provides the background for our approach;  \cref{sec:method} introduces MutexMatch, the proposed method for image classification; \cref{exp} shows the experimental results of MutexMatch on various SSL benchmark; \cref{ab_section} provides extensive ablation studies for each component in our method; \cref{sec:fd} further discusses the consistency regularization on proposed True-Negative Classifier; \cref{sec:con} concludes our contribution in this work. 

\section{Related Work}
\label{rework}
In this section, we review the related works from semi-supervised learning, consistency regularization, pseudo-labeling and complementary label.

\subsection{Semi-supervised Learning (SSL)} 
SSL aims to exploit ample unlabeled data for mitigating the lack of labeled data. In a word, we need to utilize the guidance information provided by limited labeled data to mine potential data representation of unlabeled data, 
which can be summarized as the following optimization task:
$$\min \mathcal{L} =\mathcal{L} _{s}(x_{lb},y_{lb};\theta)+\mathcal{L}_{u}(x_{ulb};\theta),$$
where $\theta$ is the parameter of model, $\mathcal{L}_{s}$ is the supervised loss for labeled data $x_{lb}$ (the corresponding label is $y_{lb}$) and $\mathcal{L}_{u}$ is the unsupervised loss for unlabeled data $x_{ulb}$.
SSL is a very promising technique in many fields of machine learning, \eg, classification \cite{sohn2020fixmatch,rizve2021in,tai2021sinkhorn,zhao2022dc,duan2022rda}, detection \cite{jeong2019consistency,sohn2020simple} and segmentation \cite{papandreou2015weakly,shi2021inconsistency,yang2022st++}, because it breaks the dilemma that labeling data consumes a lot of time and energy. 
Additionally, semi-supervised learning could be applied to a broader scene, \eg, domain adaptation \cite{chen2020semi}, deep hashing \cite{hu2021adversarial}, 
person re-identification \cite{qi2020progressive}, 
distributed learning \cite{fierimonte2016fully} and so on. In addition, generative models are also play a significant role in SSL, which including 1) semi-supervised Variational Auto-Encoder (VAE)  \cite{kulkarni2015deep,paige2017learning,li2019disentangled,joy2020rethinking}; 2) semi-supervised Generative Adversarial Net (GAN) \cite{dai2017good,salimans2016improved,qi2018global,wei2018improving}. Although generative models are often considered as unsupervised methods, they can be used to facilitate SSL since they can learn the distribution of real data from unlabeled samples, \eg, one way that is most relevant to this paper is to regularize classifiers using samples produced by GANs. In this paper, we focus on the semi-supervised learning for image classification. A similar strategy is prevalent in current semi-supervised classification work, model is trained on labeled data and pseudo-labels are generated for unlabeled data based on model's predictions \cite{lee2013pseudo,berthelot2019mixmatch,berthelot2020remixmatch,sohn2020fixmatch,li2021comatch}, which boosts the performance of SSL combining consistency regularization (described below). 

% \noindent \textbf{Consistency regularization} 
\subsection{Consistency Regularization}
Consistency regularization is a significant branch of recent state-of-the-art (SOTA) SSL methods, which is proposed in~\cite{bachman2014learning}. 
Such methods encourage the classifier to output the same class probability distribution after different versions of augmentation for the same unlabeled data. Generally speaking, the consistency regularization based models are trained with unlabeled data using the loss function: 
$\| p(y\vert \alpha (x))-p(y| \alpha (x))\|^{2}_{2},$
where $x$ is the input image, $p(\cdot)$ and $\alpha (\cdot)$ are stochastic functions \cite{bachman2014learning}. 
% is an kind of transformation that does not change the image label. 
Particularly, $\alpha (\cdot )$ can adopt different augmentation methods, \eg,
% Mixup~\cite{zhang2017mixup} in~\cite{berthelot2019mixmatch},
RandAugment~\cite{sohn2020fixmatch} in~\cite{sohn2020fixmatch} and CTAugment in~\cite{berthelot2020remixmatch}. 

Specifically, \cite{laine2016temporal} enforces a loss of consistency on the predictions of two augmented variants of unlabeled data.
In~\cite{tarvainen2017mean}, a teacher model is maintained to generate more stable targets for unlabeled data, while the mean squared error is used to encourage the same predictions of both the student and teacher models.
% In~\cite{miyato2019virtual}, adversarial attack is adopted for consistent regularization. 
\cite{xie2020unsupervised} adopts automatic augmentation for data perturbation and enforces a loss of consistency by the KL divergence.  some other models seek to improve the consistency of predicted outputs over the perturbed data to
the train semi-supervised models. The generative model mentioned above can also be applied to consistency regularization. GAN-based methods (\eg, \cite{wei2018improving,qi2020loss}) train a multi-class discriminator to distinguish fake samples from the real classes,  improving the consistency of predictions over the perturbed data.
Recently, some holistic methods~\cite{berthelot2020remixmatch,sohn2020fixmatch,li2021comatch,guiimproving} have been proposed to combine consistency regularization with pseudo-labeling for better SSL performance. 

Differently, in MutexMatch,  we propose a novel mutex-based consistency to learn an informative representation of all unlabeled data.  
Precisely, our method has a large distinction from previous methods involved consistency regularization: most previous works either 1) utilizing a predefined threshold for determining certain/uncertain predictions on unlabeled data or 2) conducting consistency regularization between predictions on different augmentations of the same image. Our goal is to utilize uncertain samples by employing consistency regularization on complementary predictions. This is a compromise to avoid the influence of noise pseudo-labels (\ie, \textit{``what it is''}) and still have the ability to learn the data representation of uncertain samples by asking \textit{``what it is not''}. Meanwhile, we propose a flexible way to use consistency regularization on complementary labels, \ie, we encourage the model to maintain a consistent degree of dissimilarity in the top-$k$ dissimilar classes.  

% \noindent \textbf{Pseudo-labeling} 
\subsection{Pseudo-labeling}
Pseudo-labeling is widely leveraged to achieve entropy minimization \cite{grandvalet2004semi} by constructing one-hot labels from predictions of unlabeled data and making use of them based on cross-entropy loss \cite{lee2013pseudo}. However, recent pseudo-labeling based methods \cite{sohn2020fixmatch,li2021comatch} all have a significant limitation, \ie, using a threshold to select high-confidence pseudo-labels results in the waste of low-confidence samples. We notice \cite{rizve2021in} proposes an uncertainty-aware pseudo-label selection framework to use both high and low-confidence samples. 
However, it introduces two thresholds to control the pseudo-label generation, leading partial samples to remain unutilized.
%thus the samples in between remain unutilized. 
Moreover, 
the way it utilizes low-confidence samples is not very informative, which can be demonstrated in \cref{sec:uls}.

% \noindent \textbf{Complementary label} 
\subsection{Complementary Label}
Complementary label \cite{ishida2017learning,ishida2018complementary,kim2019nlnl,rizve2021in} is used to help the model learn which class the input image does not belong to. Considering a $c$-class classification task, we denote $x\in \mathcal{X}$ as an input image and $y\in \mathcal{Y} =\{1,...,c\}$ as its positive label. Complementary label $\overline{y}$ is generated by selecting from $\mathcal{Y} \setminus \{y\}$ at random. Different from \cite{kim2019nlnl}, we design a novel way (detailed in \cref{tnc_section}) to propose complementary labels, so as to ensure their effectiveness in semi-supervised learning, while experiments on using standard generation of complementary label will be discussed in \cref{abl:strategy}.

\section{MutexMatch}
\label{sec:method}
\subsection{Overview}

Different from existing SSL approaches, besides a feature extractor $\theta (\cdot)$, MutexMatch jointly trains two distinct classifiers, a True-Positive Classifier (TPC) $\mathcal{P}(\cdot)$ and a True-Negative Classifier (TNC) $\mathcal{N} (\cdot)$. In specific, TPC is used to predict which class the instance belongs to (\ie, used for test phase), while TNC is employed to indicate which class the instance does not belong to (\ie, true-negative). To mitigate pseudo-labeling errors, a predefined high-confidence threshold $\tau$ is utilized to split the unlabeled data into high-confidence and low-confidence portions. 
Besides training TPC on high-confidence portion, we explore complementary labels on low-confidence samples to train TNC. 
In this way, all the unlabeled data could be effectively exploited.

We have $B$ labeled data $\mathcal{X} =\{(x^{lb}_{n},y^{lb}_{n})\}^{B}_{n=1}$ and $\mu B$ unlabeled data $\mathcal{U}=\{(x^{ulb}_{n})\}^{\mu B}_{n=1}$ in a mini-batch, where $\mu$ represents the relative size of $\mathcal{X}$ and $\mathcal{U}$. Following~\cite{sohn2020fixmatch}, we perform weak and strong augmentations for data perturbations, denoted by $\alpha_{w}(\cdot)$ and $\alpha_{s}(\cdot)$, respectively. Given weakly-augmented instances $x^{w}=\alpha_{w}(x^{ulb})$ and strongly-augmented instances $x^{s}=\alpha_{s}(x^{ulb})$, MutexMatch simultaneously optimizes four losses: the supervised loss $\mathcal{L} _{sup}$, the separated negative loss $\mathcal{L}_{sep}$, the positive consistency loss $\mathcal{L}_{p}$ and the negative consistency loss $\mathcal{L}_{n}$. To sum up, the total loss is
\begin{equation}
\mathcal{L} =\mathcal{L} _{sup}+\lambda _{sep}\mathcal{L}_{sep}+\lambda_p \mathcal{L}_{p}+\lambda _{n}\mathcal{L} _{n},
\label{eq:allloss}
\end{equation}
where  $\lambda _{sep}$, $\lambda_p$ and $\lambda _{n}$ are weight hyper-parameters to balance the relative importance of corresponding losses. The supervised loss $\mathcal{L} _{sup}$ is the cross-entropy between $y^{lb}$ and the predictions of TPC on labeled data $x^{lb}$, calculated as follows:
\begin{equation}
\label{s} 
\mathcal{L}_{sup}=\frac{1}{B}\sum_{n = 1}^{B}H(y^{lb}_{n},\mathcal{P} (\theta( \alpha_w (x^{lb}_{n}) ))),
\end{equation}
where $H(p,q)$ denotes the standard cross-entropy loss between distribution $q$ and $p$.
% \section{Algorithm}

\begin{figure}[t]
\vskip 0in
% \setlength{\abovecaptionskip}{-1mm}
%    \vspace{-1em}
   \centering
   \subfloat[]{
   \includegraphics[width=3.9cm,height=3.4cm]{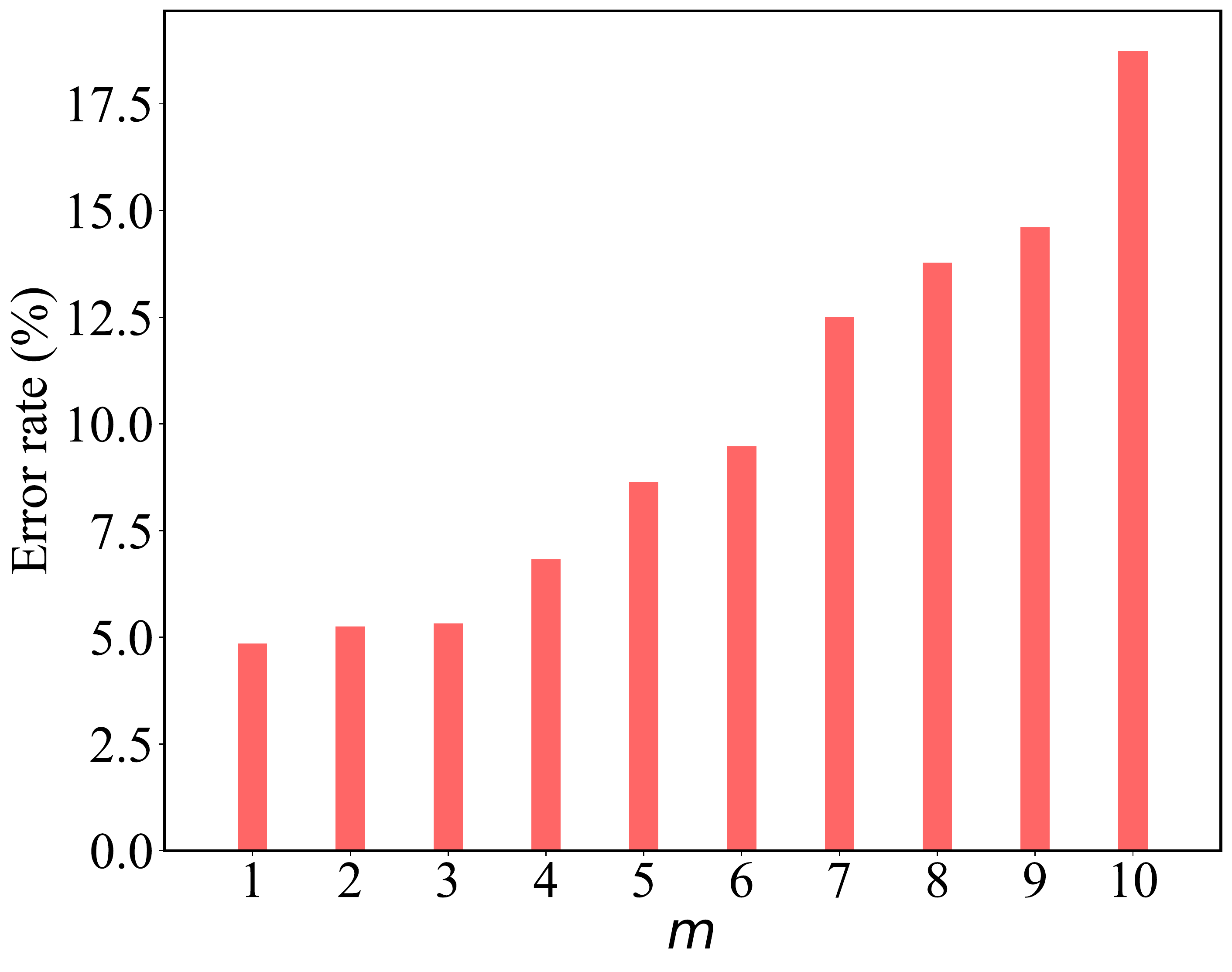}
   \label{fig:short-3a}
   }
%   \hspace{-3mm}
\hfil
   \subfloat[]{
   \includegraphics[width=3.9cm,height=3.4cm]{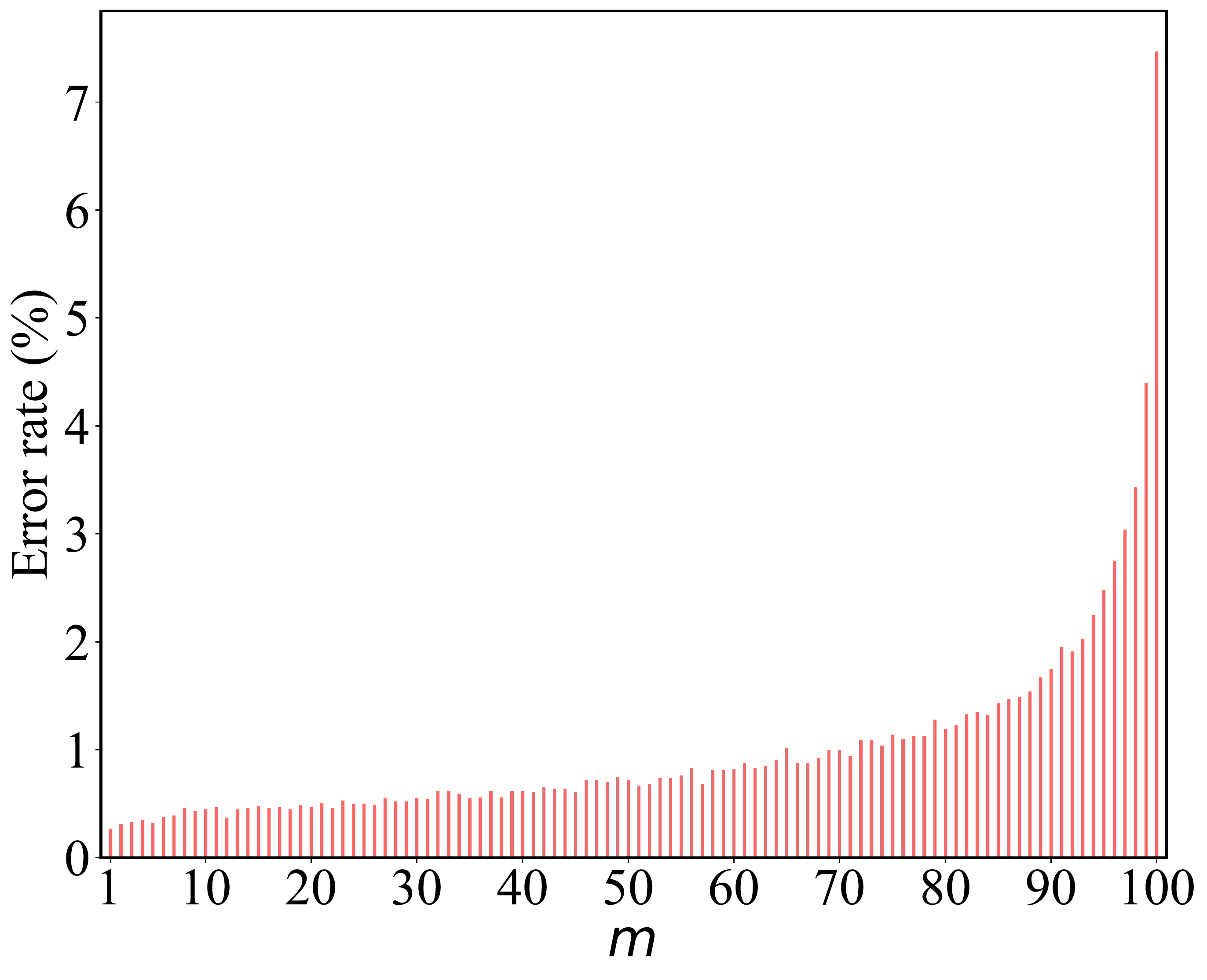}
   \label{fig:short-3b}
   }
% \vspace{-0.2in}
   \caption{Training models (WRN-28-2 \cite{zagoruyko2016wide}) on CIFAR-10/100 with only 40/400 labels are shown in (a)/(b). We show the error rate of complementary label with various $m$, where $m$ indicates that the class index with $m$-th smallest probability in the prediction of TPC is chosen to serve as the complementary label and we choose $\arg\min(\mathcal{P} (\theta (x^{w})))$ in MutexMatch.} 
   \label{fig:arg}
%   \vskip 0in
\end{figure}   
 \begin{algorithm*}[t]
      \caption{MutexMatch: Semi-supervised Learning with Mutex-based Consistency Regularization} 
      \label{alg}
    \begin{algorithmic}
         \STATE {\bfseries Input:} batch of labeled data $\mathcal{X} =\{(x^{lb}_{n},y^{lb}_{n})\}^{B}_{n=1}$, batch of unlabeled data $\mathcal{U} =\{x^{ulb}_{n}\}^{\mu B}_{n=1}$, feature extractor $\theta $, TPC $\mathcal{P} $, TNC $\mathcal{N}$, hyper-parameter $k$ for a top-$k$ algorithm
         \FOR{{iteration} $t$ }
            \STATE $\mathcal{L}_{sup}=\frac{1}{B}\sum_{n = 1}^{B}H(y^{lb}_{n},\mathcal{P} (x^{lb}_{n} ))$\hfill\quad\textcolor{light-gray}{$//$ \texttt{Supervised loss for $x^{lb}$}}
            \FOR{{iteration} $n=1$ {\bfseries to} $\mu B$}
                 \STATE $p^{w}_{n}=\hat{\mathcal{P}} (\hat{\theta} (\alpha _{w}(x^{ulb}_{n})))$\hfill\quad\textcolor{light-gray}{$//$ \texttt{Compute TPC's prediction for weakly-augmented $x^{ulb}$}}
                 \STATE $p^{s}_{n}=\mathcal{P} (\theta (\alpha _{s}(x^{ulb}_{n})))$\hfill\quad\textcolor{light-gray}{$//$ \texttt{Compute TPC's prediction for strongly-augmented $x^{ulb}$}}
                 \STATE $r^{w}_{n}=\hat{\mathcal{N}} (\hat{\theta }(\alpha _{w}(x^{ulb}_{n})))$\hfill\quad\textcolor{light-gray}{$//$ \texttt{Compute TNC's prediction for weakly-augmented  $x^{ulb}$}}
                 \STATE $g_{n,(i)}=\mathbbm{1}(r^{w}_{n,(i)} \in \mathcal{T}_{k}(r_{n}^{w}))$\hfill\quad\textcolor{light-gray}{$//$ \texttt{Control the intensity of consistency regularization on $\mathcal{N}$}}
                 \STATE $r^{s}_{n}=\mathcal{N} (\theta (\alpha _{s}(x^{ulb}_{n})))$\hfill\quad\textcolor{light-gray}{$//$ \texttt{Compute TNC's prediction for strongly-augmented $x^{ulb}$}}
                 \STATE $\hat{p}^{w}_{n}= \arg \max (p^{w}_{n})$\hfill\quad\textcolor{light-gray}{$//$ \texttt{Select pseudo-label for $x^{ulb}$}}
                 \STATE $\hat{q}^{w}_{n}= \arg \min (p^{w}_{n})$\hfill\quad\textcolor{light-gray}{$//$ \texttt{Select complementary pseudo-label for $x^{ulb}$}}
             \ENDFOR
             
         \ENDFOR
         \STATE $\mathcal{L}_{sep}=\frac{1}{\mu B}\sum_{n = 1}^{\mu B}H(\hat{q}^{w}_{n},\mathcal{N} (\hat{\theta} (x^{w}_{n}) ))$\hfill\quad\textcolor{light-gray}{$//$ \texttt{Stop back-propagating gradient on $\theta$}}
         \STATE $\mathcal{L}_{p}=\frac{1}{\mu B}\sum_{n = 1}^{\mu B}\mathbbm{1}(\max (p^{w}_{n})\geq  \tau )H(\hat{p}^{w}_{n},p^{s}_{n})$\hfill\quad\textcolor{light-gray}{$//$ \texttt{Positive consistency loss for $x^{ulb}$}}
         \STATE $\mathcal{L}_{n}=\frac{1}{\mu B}\sum_{n = 1}^{\mu B}\mathbbm{1}(\max (p^{w}_{n})<\tau )(-\frac{1}{k}\sum^{C}_{i=1}g_{n,(i)}r_{n,(i)}^{w}\log(r_{n,(i)}^{s}))$\hfill\quad\textcolor{light-gray}{$//$ \texttt{Negative consistency loss for $x^{ulb}$}}
         \STATE  update $\theta ,\mathcal{P} , \mathcal{N} $ by SGD to optimise $\mathcal{L}_{sup}+\lambda _{sep}\mathcal{L}_{sep}+\lambda _{p}\mathcal{L}_{p}+\lambda _{n}\mathcal{L}_{n} $
    \end{algorithmic}
  \end{algorithm*}
\subsection{True-Negative Classifier}
\label{tnc_section}
In multi-class classification tasks, it is easier to predict ``what it is not'' than to decide exactly ``what it is''. For example, given an image of airplane in CIFAR-10, we can predict which class it does not belong to with a 90\% probability  at random, whereas there is merely a 10\% probability  to correctly predict it indeed is an airplane. To this end, we design a True-Negative Classifier to predict what it is not. Considering that compared to TPC, it is much easier to obtain correct labels for TNC, 
%Thus for unlabelled data, though their exact classes are unknown, we can still obtain useful information from knowing what they are not. 
we first exploit TNC to provide more guidance information on unlabeled data. 
% We then propose mutex-based consistency regularization on both TPC and TNC to make full use of unlabeled data in next section. 
% The high-level training process of TNC is shown in \cref{tnc_fig}. 

Unlike the standard way of generating complementary label~\cite{ishida2017learning,yu2018learning}, we use the class with the lowest confidence in TPC's prediction as the complementary label to train TNC, which is a simple strategy to obtain a robust TNC. Intuitively, using the class with the lowest probability in the prediction is less likely to be the correct class of the image, \ie, the complementary label is more likely to be correct. This claim can be demonstrated in \cref{fig:arg}.
The separate training loss $\mathcal{L}_{sep}$ for TNC can be calculated as
\begin{equation}
\mathcal{L}_{sep}=\frac{1}{\mu B}\sum_{n = 1}^{\mu B}H(\arg \min (\hat{\mathcal{P}} (\hat{\theta} (x^{w}_{n})),\mathcal{N} (\hat{\theta} (x^{w}_{n}) ))), 
 \label{sep} 
\end{equation}
where $\hat{x}$ represents we stop back-propagating gradient on $x$. Since our downstream task is to accurately classify images, 
we adopt such gradient-blocking operation to ensure that the feature extractor will not be affected by the training of TNC (because it predicts complementary label). Furthermore, we extensively investigate the effectiveness of TNC in \cref{sec:atnc}.

\nobalance

\subsection{Mutex-based Consistency Regularization}
\label{cos_section}
In recent consistency-regularization based SSL methods, only samples with high-confidence predictions are leveraged to train models, which  could lead to inefficient utilization of unlabeled data, especially at the early stage of the training process. By contrast, MutexMatch can also learn an informative representation of low-confidence unlabeled samples via introducing a novel mutex-based consistency regularization. A high-confidence threshold $\tau$ on TPC's predictions is defined to separate the unlabeled samples into two portions with mutex confidence intervals, \ie, the high-confidence one ($\geq \tau$) and the low-confidence one ($<\tau$). In the high-confidence portion, we use TPC to learn what the unlabeled data is, while in the low-confidence portion, we employ TNC to learn what it is not, as it is difficult for us to obtain its real class information.

On the one hand, we first use weakly-augmented example $x^{w}$ to generate pseudo-labels from TPC and enforce positive consistency against its corresponding strongly-augmented variant $x^{s}$. Then we obtain their predictions, $p^{w}=\hat{\mathcal{P}} (\hat{\theta}(x^{w}))$ and $p^{s}=\mathcal{P} (\theta (x^{s}))$.
Let $\hat{p}^{w}=\arg \max (p^{w})$, such consistency can be achieved by minimizing the positive consistency loss $\mathcal{L}_{p}$: 
\begin{equation}
\mathcal{L}_{p}=\frac{1}{\mu B}\sum_{n = 1}^{\mu B}\mathbbm{1}(\max (p^{w}_{n})\geq  \tau )H(\hat{p}^{w}_{n},p^{s}_{n}),  
\label{p} 
\end{equation}
where $\mathbbm{1}(\max (p^{w})> \tau)$ retains the predictions whose maximum probabilities are larger than $\tau$.  
For entropy minimization~\cite{lee2013pseudo}, we adopt hard pseudo-label $\hat{p}^{w}$ to enforce the consistency regularization on TPC.

On the other hand, for the augmented images $x^{w}$ and $x^{s}$ of these low-confidence samples, TNC computes predictions $r^{w}=\hat{\mathcal{N}} (\hat{\theta} (x^{w}))$ and $r^{s}=\mathcal{N} (\theta (x^{s}))$ for them. We simply treat the probability component $r^{w}_{(i)}$ in $r^{w}$ as the degree of dissimilarity from $i$-th class, where $i\in(1,\dots,C)$ and $C$ is the total amount  of classes. We enforce consistency regularization on the degree of dissimilarity, which is achieved by minimizing the negative consistency loss $\mathcal{L}_{n}$:
\begin{equation}
\mathcal{L}_{n}=\frac{1}{\mu B}\sum_{n = 1}^{\mu B}\mathbbm{1}(\max (p^{w}_{n})<  \tau )(-\frac{1}{k}\sum^{C}_{i=1}g_{n,(i)}r_{n,(i)}^{w}\log(r_{n,(i)}^{s})),   
\label{eq:ln}
\end{equation}
where $k$ is the hyper-parameter used to control the intensity of consistency. 
The $g_{n}$ is a $C$-dimensional binary vector indicating the selected components in prediction $r_{n}^{w}$, which is defined as: $g_{n,(i)}=\mathbbm{1}(r^{w}_{n,(i)} \in \mathcal{T}_{k}(r_{n}^{w}))$. $\mathcal{T}_{k}$ represents the top-$k$ algorithm used to select the top $k$ largest probability components in prediction. When $k=C$, we use $H(r^{w}_{n},r^{s}_{n})$ to replace term $-\frac{1}{k}\sum^{C}_{i=1}g_{n,(i)}r_{n,(i)}^{w}\log(r_{n,(i)}^{s})$ in \cref{eq:ln} for simplicity (\ie, all components in $r^{w}_{n}$ are selected).
We use soft pseudo-label $r^{w}$ for consistency regularization on TNC, \ie, 
we encourage TNC to maintain consistent prediction probability distributions for different augmented versions of the same image, 
while the setting of consistency using soft-label will be discussed  in \cref{abl:strategy}. The pseudo-code of proposed MutexMatch is presented in  \cref{alg}.

\subsection{Theoretical Analysis}
Following \cite{wei2020theoretical}, we study MutexMatch and prove a lower error bound than conventional consistency-based methods without TNC, \eg, current SOTA SSL method FixMatch \cite{sohn2020fixmatch}. 
% We define a population objective which measures positive and negative consistency and accuracy of pseudo-label. Holding Assumptions 3.3 and 4.1 in  \cite{wei2020theoretical}, 
We show that minimizing both positive consistency loss and negative consistency loss in a mutex-based way leads to high accuracy on pseudo-labels, which in turn boosts the performance of SSL.

Firstly, we denote $P$ as the distribution of unlabeled samples $x^{ulb}$ over input space  $\mathcal{U}$. We let $G_{pl}$ denote a classifier trained on labeled data $x^{lb}\in\mathcal{X}$.
Next, we define the population losses of positive consistency (for positive label) and negative consistency (for complementary label) respectively. Given transformation $\alpha$, following \cite{wei2020theoretical}, we define the population positive consistency loss as:
% \begin{equation}
%     \mathcal{R}_{\alpha}(\mathcal{P})=\mathbbm{E}_{P}[\mathbbm{1}(x: \mathcal{P}(x^{w})\neq\mathcal{P}(x^{s}))].
%     \label{eq:ppcl}
% \end{equation}
\begin{equation}
    \mathcal{R}_{\alpha}(\mathcal{P})=\mathbbm{E}_{P}[\mathbbm{1}(x: p^{w}\neq p^{s})].
    \label{eq:ppcl}
\end{equation}
Considering that a class has multiple complementary labels, we can't directly define the population negative  consistency loss as Equation \eqref{eq:ppcl} . Thus, 
% denoting  $\mathcal{T}_{k}(x)=\mathcal{T}_{k}(\mathcal{N}(x))$, 
we consider using a similarity metric function 
$\xi(\mathcal{T}_{k}(r^{w}),\mathcal{T}_{k}(r^{s}))$, \eg, cross-entropy, 
% $\xi=\frac{|\mathcal{T}_{k}(\alpha_{w}(x))
%     \cap \mathcal{T}_{k}(\alpha_{s}(x))|}{|\mathcal{T}_{k}(\alpha_{w}(x))
%     \cup \mathcal{T}_{k}(\alpha_{s}(x))|}$ 
to define the population negative  consistency loss:
\begin{equation}
    \mathcal{R}_{\alpha}(\mathcal{N})=\mathbbm{E}_{P}[\mathbbm{1}(x: \xi(\mathcal{T}_{k}(r^{w}),\mathcal{T}_{k}(r^{s}))\leq  t)],
\end{equation}
where $t$ is a threshold used to determine whether $r^{w}$ and $r^{s}$ (\ie, predictions of TNC) are consistent. Choosing an appropriate  $t$ can enable us to obtain a reliable estimation for negative consistency loss $\mathcal{L}_{n}$ that uses soft complementary labels. 
% $\mathcal{R}_{\alpha}(\mathcal{N})\approx \frac{\mathcal{R}_{\alpha}(\mathcal{P})}{\mathcal{L}_{p}}\mathcal{L}_{n}$.
Next, we introduce the following two reasonable assumptions to help investigate our analysis.

We require the following assumption that it is easier for the classifier to achieve consistency on complementary pseudo-labels than that on pseudo-labels. For a classifier that has not been implemented with any consistency regularization, it is very difficult to keep consistent predictions on different augmented versions of the same image. However,  it is simpler to achieve a tolerant consistency on a set of complementary labels, \ie, as long as there is  a partial overlap in the degree of dissimilarity as described in \cref{sec:intro}.
\begin{assumption}
We assume  $P$ satisfies \textit{Assumption 4.1} in \cite{wei2020theoretical} for some \textit{expansion factor} $c$, and satisfies \textit{Separation Assumption 3.3} in \cite{wei2020theoretical} for some $\mu$ and $\omega$. For positive label,  $P$ is $\alpha$-separated with probability $1-\mu$ by ground-truth classifier $G^{\star}$, as follows: $R_{\mathcal{\alpha}}(G^{\star})\leq \mu$.  For complementary label,  $P$ is $\alpha$-separated with probability $1-\omega$ by complementary ground-truth classifier $\overline{G^{\star}}$, as follows: $R_{\mathcal{\alpha}}(\overline{G^{\star}})\leq \omega$. And we suppose  $\omega<\frac{\mu}{2}$, which is based on our claim that consistency on  complementary labels is much easier to achieve.
\label{ass:sep}
\end{assumption}

We define robust set of $\mathcal{P}$: $\mathcal{S}_{\alpha}(\mathcal{P})=	\left\{x:p^{w}=p^{s}	\right\}$, and robust set of $\mathcal{N}$:   $\mathcal{S}_{\alpha}(\mathcal{N})=\left\{x:\xi(\mathcal{T}_{k}(r^{w}),\mathcal{T}_{k}(r^{s}))> t\right\}$, which means $\mathcal{P}$ and $\mathcal{N}$ are robust for images under transformation $\alpha$ respectively. Now we introduce our key assumption as follows: there are few samples are capable of achieving prediction consistency on both pseudo-labels and complementary labels. 
\begin{assumption}
We assume  $P(\mathcal{S}_{\alpha}(\mathcal{P})\setminus \mathcal{S}_{\alpha}(\mathcal{N}))\geq P(\mathcal{S}_{\alpha}(\mathcal{P})\cap \mathcal{S}_{\alpha}(\mathcal{N}))$.
\label{ass:set}
\end{assumption}
In MutexMatch, we perform mutex-based consistency regularization on disjoint sets of samples (\ie, high-confidence portion and low-confidence portion) indicating the right-hand side of \cref{ass:set} is relatively small. Thus, it is reasonable to believe this assumption always holds.

% Here we introduce our key assumption: it is easier for the classifier to achieve consistency on complementary pseudo-labels than that on pseudo-labels. For a classifier that has not been implemented with any consistency regularization, it is very difficult to predict different augmented versions of the same image as the same class. However, 
% for these different augmented versions, it is simpler to achieve a tolerant consistency on a set of ``what it is not'', \ie, as long as there is  a partial overlap in the degree of dissimilarity as described in \cref{sec:intro}.
% \begin{assumption}
% We assume  $P$ satisfy the \textit{Separation Assumption 3.3} in \cite{wei2020theoretical} for some $\mu$ and $\omega$. For  positive labels,  $P$ is $\alpha$-separated with probability $1-\mu$ by ground-truth classifier $G^{\star}$, as follows: $R_{\mathcal{\alpha}}(G)\leq \mu$.  For complementary labels,  $P$ is $\alpha$-separated with probability $1-\omega$ by complementary ground-truth classifier $\overline{G^{\star}}$, as follows: $R_{\mathcal{\alpha}}(\overline{G^{\star}})\leq \omega$. And we suppose  $\omega<\frac{\mu}{2}$, which is based on our claim that consistency on  complementary labels is easier to achieve.
% \label{ass:sep}
% \end{assumption}
Given a classifier $\mathcal{F}$ of conventional consistency-based model (\eg, FixMatch), we design the following objective over $\mathcal{F}$:
$
    \mathop{\min}_{\mathcal{F}}\mathcal{L}(\mathcal{F})=\frac{c+1}{c-1}L_{0\text{-}1}(\mathcal{F},G_{pl})+\frac{2c}{c-1}R_{\alpha}(\mathcal{F}) -Err(G_{pl}),
    % \label{eq:fix}
$
where $L_{0\text{-}1}(G,G')=\mathbbm{E}_{P}[\mathbbm{1}(G(x)\neq G'(x))]$ is defined to be the disagreement between  $G$ and $G'$. According to Lemma 4.2 in \cite{wei2020theoretical}, we have $Err(\mathcal{F})\leq \mathcal{L}(\mathcal{F})$. For any minimizer $\widehat{\mathcal{F}}$ of $\mathop{\min}_{\mathcal{F}}\mathcal{L}(\mathcal{F})$, we denote its error bound as $B_{\mathcal{F}}$.
% $Err(\widehat{\mathcal{F}})\leq B_{\mathcal{F}}$.
Given  $\mathcal{P}$ and  $\mathcal{N}$ in MutexMatch,
denoting $L_{0\text{-}1}(\mathcal{P},G_{pl})+R_{\alpha}(\mathcal{P})- (c-1)Err(G_{pl})$ as $\varphi$, 
we design the following objective over $\mathcal{P}$:
\begin{align}
% \scriptsize  
% \fontsize{7.6pt}{\baselineskip}\selectfont
    &\mathop{\min}_{\mathcal{P}}\mathcal{L}(\mathcal{P})=\nonumber \\
& \quad \quad\quad \begin{cases}
\frac{2}{c-1}L_{0\text{-}1}(\mathcal{P},G_{pl})+\frac{c+1}{c-1}R_{\alpha}(\mathcal{P})+2R_{\alpha}(\mathcal{N})& \text{$\varphi\leq 0$}\\
\frac{c+1}{c-1}L_{0\text{-}1}(\mathcal{P},G_{pl})+\frac{2c}{c-1}R_{\alpha}(\mathcal{P})-Err(G_{pl})& \text{$\varphi>0$}
\end{cases}.
\label{eq:mutex}
\end{align}

For any minimizer $\widehat{\mathcal{P}}$ of \cref{eq:mutex}, we denote its error bound as $B_{\mathcal{P}}$. 
The next theorem shows that the error bound of SSL methods with mutex-based consistency is lower than that of conventional consistency-based methods.
% $Err(\widehat{\mathcal{P}})\leq B_{\mathcal{P}}$.
\begin{theorem}
\label{the}
% Suppose \cref{ass:sep} and \ref{ass:set}  hold.  
% Given a classifier $\mathcal{F}$ of FixMatch, in the setting of the Theorem 4.3 in \cite{wei2020theoretical}, $\mathcal{F}$ satisfies the following error bound:
% \begin{equation}
%     Err(\mathcal{F})\leq B_{\mathcal{F}}=\frac{2}{c-1}Err(G_{pl})+\frac{2c}{c-1}\mu.
% \end{equation}
% Given TPC $\mathcal{P}$ in MutexMatch,
% denoting $L_{0\text{-}1}(\mathcal{P},G_{pl})+R_{\alpha}(\mathcal{P})- (c-1)Err(G_{pl})$ as $\varphi$, where $L_{0\text{-}1}(\mathcal{P},G_{pl})=\mathbbm{E}_{P}[\mathbbm{1}(\mathcal{P}(x)\neq G_{pl}(x))]$ is defined to be the disagreement between  $\mathcal{P}$ and $G_{pl}$,
% $\mathcal{P}$ satisfies the following error bound:
% \begin{equation}
%     Err(\mathcal{P})\leq B_{\mathcal{P}}=
% \begin{cases}
% \frac{2}{c-1}Err(G_{pl})+\frac{c+1}{c-1}\mu+2\omega& \text{$\varphi\leq 0$}\\
% B_{\mathcal{F}}& \text{$\varphi>0$}
% \end{cases}.
% \end{equation}
% If $\omega<\frac{\mu}{2}$ (\cref{ass:sep}), we have $B_{\mathcal{P}}\leq B_{\mathcal{F}}$ immediately.
Suppose \cref{ass:sep} and \ref{ass:set} hold, we have $Err(\mathcal{P})\leq\mathcal{L}(\mathcal{P})$ and  $B_{\mathcal{P}}\leq B_{\mathcal{F}}$.
\end{theorem}

To prove this theorem, we give the proof of the error bound of TPC $\mathcal{P}$ in MutexMatch firstly. Then \cref{the} follows immediately by holding \cref{ass:sep}.
The proofs follow the analysis of \cite{wei2020theoretical}. We first give the main notations of \cite{wei2020theoretical} which are used in following proofs. 
$\mathcal{M}(\mathcal{P})=\left\{x:\mathcal{P}(x)\neq G^{\star}(x)\right\}$ denotes the set of examples pseudolabeled mistakenly by $\mathcal{P}$, where $G^{\star}(x)$ is the ground-truth of unlabeled sample $x$.
Following Theorem A.2 in \cite{wei2020theoretical}, we define three disjoint sets of $\mathcal{M}(\mathcal{P})\cap\mathcal{S}_{\alpha}(\mathcal{P})$: $$\mathcal{M}_{1}=\left\{{x:\mathcal{P}(x)=G_{pl}(x),G_{pl}(x)\neq G^{\star}(x),x\in\mathcal{S}_{\alpha}(\mathcal{P})}\right\},$$
\begin{align}
    \mathcal{M}_{2}=\{x:&\mathcal{P}(x)\neq G_{pl}(x),G_{pl}(x)\neq G^{\star}(x),\nonumber \\  
&  G(x)\neq G^{\star}(x),x\in\mathcal{S}_{\alpha}(\mathcal{P})\},\nonumber
\end{align}
$$\mathcal{M}_{3}= \left\{{x:\mathcal{P}(x)\neq G_{pl}(x),G_{pl}(x)= G^{\star}(x),x\in\mathcal{S}_{\alpha}(\mathcal{P})}\right\}.$$
$q$ is a factor defined in Definition A.1 in \cite{wei2020theoretical} and we have $P(\mathcal{M}_{1}\cup\mathcal{M}_{2})\leq q$  by Lemma A.3 in \cite{wei2020theoretical}; $\beta$ is a factor defined in Lemma A.7 in \cite{wei2020theoretical} for choice of $q$; $c$ is the \textit{expansion factor} defined in Definition 3.1 in \cite{wei2020theoretical}. 
Moreover, given $\mathcal{F}$, a classifier of conventional consistency-based method, and the objective over  $\mathcal{F}$:
$$
    \mathop{\min}_{\mathcal{F}}\mathcal{L}(\mathcal{F})=\frac{c+1}{c-1}L_{0\text{-}1}(\mathcal{F},G_{pl})+\frac{2c}{c-1}R_{\alpha}(\mathcal{F}) -Err(G_{pl}),
    % \label{eq:fix}
$$
we have $Err(\mathcal{F})\leq\mathcal{L}(\mathcal{F})$ by Lemma 4.2 in \cite{wei2020theoretical}. And in the setting of Theorem 4.3 in \cite{wei2020theoretical}, for any minimizer $\widehat{\mathcal{F}}$ of $\mathop{\min}_{\mathcal{F}}\mathcal{L}(\mathcal{F})$, we have its error bound: $Err(\widehat{\mathcal{F}})\leq B_{\mathcal{F}}= \frac{2}{c-1}Err(G_{pl})+\frac{2c}{c-1}\mu$. 
\begin{lemma}
\label{lem}
In the setting of Theorem A.2 in \cite{wei2020theoretical}, we have
$$
    P(\mathcal{M}_{3}\cap\overline{\mathcal{S}_{\alpha}(\mathcal{N} )})\leq q+R_{\alpha}(\mathcal{N}).
$$
\begin{proof}
Following Equations (A.2) and (A.3) in \cite{wei2020theoretical}, we know $\mathcal{M}_{3}\cup\left\{x:P(x)=G_{pl}(x),x\in\mathcal{S}_{\alpha}(\mathcal{N})\right\}$ and $\mathcal{M}_{1}\cup\left\{x:G_{pl}(x)=G^{\star}(x),x\in\mathcal{S}_{\alpha}(\mathcal{N})\right\}$ are unions of disjoint sets. Thus we have

\begin{align} 
     &P(\mathcal{M}_{3}\cap\overline{\mathcal{S}_{\alpha}(\mathcal{N})})+\nonumber\\
     &\quad P(\left\{x:P(x)=G_{pl}(x),x\in\mathcal{S}_{\alpha}(\mathcal{N})\right\}\cap\overline{\mathcal{S}_{\alpha}(\mathcal{N})}) \nonumber\\
      &=P(\mathcal{M}_{1}\cap\overline{\mathcal{S}_{\alpha}(\mathcal{N})})+\nonumber\\
      &\quad P(\left\{x:G_{pl}(x)=G^{\star}(x),x\in\mathcal{S}_{\alpha}(\mathcal{N})\right\}\cap\overline{\mathcal{S}_{\alpha}(\mathcal{N})}),\nonumber
\end{align}
% $$P(\mathcal{M}_{3}\cap\overline{\mathcal{S}_{\alpha}(P)})+P(\left\{x:P(x)=G_{pl}(x),x\in\mathcal{S}_{\alpha(P)}\right\}\cap\overline{\mathcal{S}_{\alpha}(P)})=P(\mathcal{M}_{1}\cap\overline{\mathcal{S}_{\alpha}(P)})+P(\left\{x:G_{pl}(x)=G^{\star}(x),x\in\mathcal{S}_{\alpha}(P)\right\}\cap\overline{\mathcal{S}_{\alpha}(P)}),$$
rearranging we obtain
\begin{align}
     &P(\mathcal{M}_{3}\cap\overline{\mathcal{S}_{\alpha}(\mathcal{N})}) = P(\mathcal{M}_{1}\cap\overline{\mathcal{S}_{\alpha}(\mathcal{N})})+\nonumber\\
     &\quad P(\left\{x:G_{pl}(x)=G^{\star}(x),x\in\mathcal{S}_{\alpha}(\mathcal{N})\right\}\cap\overline{\mathcal{S}_{\alpha}(\mathcal{N})})- \nonumber\\
      & \quad P(\left\{x:P(x)=G_{pl}(x),x\in\mathcal{S}_{\alpha(\mathcal{N})}\right\}\cap\overline{\mathcal{S}_{\alpha}(\mathcal{N})})\nonumber\\
      & \leq P(\mathcal{M}_{1}\cap\overline{\mathcal{S}_{\alpha}(\mathcal{N})})+\nonumber\\
      &\quad P(\left\{x:G_{pl}(x)=G^{\star}(x),x\in\mathcal{S}_{\alpha}(\mathcal{N})\right\}\cap\overline{\mathcal{S}_{\alpha}(\mathcal{N})})\nonumber\\
      & \leq P(\mathcal{M}_{1})+P(\overline{\mathcal{S}_{\alpha}(\mathcal{N})})\nonumber\\
      & =  P(\mathcal{M}_{1})+R_{\alpha}(\mathcal{N})\nonumber\\
      & \leq  q+R_{\alpha}(\mathcal{N}). &\tag{using $P(\mathcal{M}_{1}\cup \mathcal{M}_{2})\leq q$} \nonumber
\end{align}
\end{proof}
\end{lemma}
Now we complete the proof of \cref{the} by combining the \cref{lem}, Assumption \ref{ass:sep} and Assumption \ref{ass:set}.
\begin{proof}[Proof of \cref{the}]
Given three aforementioned disjoint subsets of  $\mathcal{M}(\mathcal{P})\cap\mathcal{S}_{\alpha}(\mathcal{P})$: $\mathcal{M}_{1}$, $\mathcal{M}_{2}$ and $\mathcal{M}_{3}$, we know $P(\mathcal{M}(\mathcal{P})\cap \mathcal{S}_{\alpha}(\mathcal{P}))=P(\mathcal{M}_{1})+P(\mathcal{M}_{2})+P(\mathcal{M}_{3})$.  Thus, we obtain
\begin{align}
    &P(\mathcal{M}(\mathcal{P})\cap \mathcal{S}_{\alpha}(\mathcal{P})\cap\overline{\mathcal{S}_{\alpha}(\mathcal{N})})=P(\mathcal{M}_{1}\cap\overline{\mathcal{S}_{\alpha}(\mathcal{N})})+\nonumber \\
    & \quad P(\mathcal{M}_{2}\cap\overline{\mathcal{S}_{\alpha}(\mathcal{N})})+P(\mathcal{M}_{3}\cap\overline{\mathcal{S}_{\alpha}(\mathcal{N})}).
    \label{eq:13}
\end{align}
Then, holding Assumption \ref{ass:set} which is tightly related to mutex-based consistency regularization, we compute
\begin{align}
     &Err(\mathcal{P})=P(\mathcal{M(\mathcal{P})}) \nonumber \\
     &\leq P(\mathcal{M}(\mathcal{P})\cap \mathcal{S}_{\alpha}(\mathcal{P})\cap\overline{\mathcal{S}_{\alpha}(\mathcal{N})})+P(\overline{\mathcal{S}_{\alpha}(\mathcal{P})\cap\overline{\mathcal{S}_{\alpha}(\mathcal{N})}}) \nonumber\\
      & \leq P(\mathcal{M}_{1})+P(\mathcal{M}_{2})+\nonumber \\
      &\quad P(\mathcal{M}_{3}\cap\overline{\mathcal{S}_{\alpha}(\mathcal{N})})+P(\overline{\mathcal{S}_{\alpha}(\mathcal{P})\cap\overline{\mathcal{S}_{\alpha}(\mathcal{N})}})&\tag{by Equation \eqref{eq:13}}\nonumber\\
      & \leq P(\mathcal{M}_{1})+P(\mathcal{M}_{2})+q+R_{\alpha}(\mathcal{N})+P(\overline{\mathcal{S}_{\alpha}(\mathcal{P})\cap\overline{\mathcal{S}_{\alpha}(\mathcal{N})}}) &\tag{by \cref{lem}} \nonumber\\
      & \leq 2q+R_{\alpha}(\mathcal{N})+P(\overline{\mathcal{S}_{\alpha}(\mathcal{P})\cap\overline{\mathcal{S}_{\alpha}(\mathcal{N})}}) &\tag{using $P(\mathcal{M}_{1}\cup \mathcal{M}_{2})\leq q$} \nonumber\\
      & \leq 2q+R_{\alpha}(\mathcal{N})+P(\overline{\mathcal{S}_{\alpha}(\mathcal{P})\cap\mathcal{S}_{\alpha}(\mathcal{N})}) &\tag{by \cref{ass:set}} \nonumber\\
      & =   2q+R_{\alpha}(\mathcal{N})+P(\overline{\mathcal{S}_{\alpha}(\mathcal{P})}\cup \overline{\mathcal{S}_{\alpha}(\mathcal{N})})  \nonumber\\
      & \leq   2q+R_{\alpha}(\mathcal{N})+P(\overline{\mathcal{S}_{\alpha}(\mathcal{P})})+ P(\overline{\mathcal{S}_{\alpha}(\mathcal{N})}) \nonumber\\
      & =   2q+R_{\alpha}(\mathcal{N})+R_{\alpha}(\mathcal{P})+R_{\alpha}(\mathcal{N}) \nonumber\\
      & \leq   2q+R_{\alpha}(\mathcal{P})+2R_{\alpha}(\mathcal{N}).\label{eq:qq} 
\end{align}

Following the proof of Lemma A.8 in \cite{wei2020theoretical}, we prove the class-conditional variant of $Err(\mathcal{P})\leq\mathcal{L}(\mathcal{P})$ in \cref{the}. We denote the class index as $i$, where $i\in(1,\dots,C)$.  Considering the case where $L^{(i)}_{0\text{-}1}(\mathcal{P},G_{pl})+R^{(i)}_{\alpha}(\mathcal{P})\leq (c-1)Err_{i}(G_{pl})$ firstly, we have $q=\frac{\beta}{c-1}Err_{i}(G_{pl})$ from Equation A.8 in \cite{wei2020theoretical}, and $\beta=\frac{L^{(i)}_{0\text{-}1}(\mathcal{P},G_{pl})+R^{(i)}_{\alpha}(\mathcal{P})}{Err_{i}(G_{pl})}$ from Equation A.7 in \cite{wei2020theoretical}. By Equation \eqref{eq:qq}, we obtain
\begin{align}
     &Err_{i}(\mathcal{P})\leq   2q+R^{(i)}_{\alpha}(\mathcal{P})+2R^{(i)}_{\alpha}(\mathcal{N}) \nonumber\\
      &= \frac{2\beta}{c-1}Err_{i}(G_{pl})+R^{(i)}_{\alpha}(\mathcal{P})+2R^{(i)}_{\alpha}(\mathcal{N}) \tag{using $q=\frac{\beta}{c-1}Err_{i}(G_{pl})$}\nonumber\\ 
      &= \frac{2(L^{(i)}_{0\text{-}1}(\mathcal{P},G_{pl})+R^{(i)}_{\alpha}(\mathcal{P}))}{c-1}+R^{(i)}_{\alpha}(\mathcal{P})+2R^{(i)}_{\alpha}(\mathcal{N}) \tag{using $\beta=\frac{L^{(i)}_{0\text{-}1}(\mathcal{P},G_{pl})+R^{(i)}_{\alpha}(\mathcal{P})}{Err_{i}(G_{pl})}$}\nonumber\\ 
      &= \frac{2}{c-1}L^{(i)}_{0\text{-}1}(\mathcal{P},G_{pl})+\frac{c+1}{c-1}R^{(i)}_{\alpha}(\mathcal{P})+2R^{(i)}_{\alpha}(\mathcal{N})=\mathcal{L}_{i}(\mathcal{P}).\label{eq:case1}
    %   &\leq \frac{2}{c-1}Err(G_{pl})+\frac{c+1}{c-1}\mu+2\omega &\text{(by \cref{ass:sep})}\nonumber\\
    %   &=B_{\mathcal{P}}\leq \frac{2}{c-1}Err(G_{pl})+\frac{2c}{c-1}\mu=B_{\mathcal{F}}. &\text{(using $\omega\leq\frac{1}{2}\mu$ in \cref{ass:sep})}\nonumber
\end{align}
Considering the case where $L^{(i)}_{0\text{-}1}(\mathcal{P},G_{pl})+R^{(i)}_{\alpha}(\mathcal{P})> (c-1)Err_{i}(G_{pl}),$ according to Equations (A.13)$\sim$(A.17) in \cite{wei2020theoretical}, we have
\begin{align}
    Err_{i}(\mathcal{P})&\leq  \frac{c+1}{c-1}L^{(i)}_{0\text{-}1}(\mathcal{P},G_{pl})+\frac{2c}{c-1}R^{(i)}_{\alpha}(\mathcal{P})-Err_{i}(G_{pl})\nonumber \\
    &=\mathcal{L}_{i}(\mathcal{P}).\label{eq:case2}
\end{align}
Then $Err(\mathcal{P})\leq\mathcal{L}(\mathcal{P})$ follows simply by taking the expectation of $Err_{i}(\mathcal{P})\leq\mathcal{L}_{i}(\mathcal{P})$ over all the classes.

Following the proof of Lemma 4.2 in \cite{wei2020theoretical}, noting that $G^{\star}$ satisfies $L_{0\text{-}1}(G^{\star},G_{pl})=Err(G_{pl})$ and  $R_{\alpha}(G^{\star})\leq \mu$, $R_{\alpha}(\overline{G^{\star}})\leq \omega$ by \cref{ass:sep}, for any minimizer $\widehat{\mathcal{P}}$ of \cref{eq:mutex}, in the case where $L_{0\text{-}1}(\mathcal{P},G_{pl})+R_{\alpha}(\mathcal{P})\leq (c-1)Err(G_{pl})$, by \cref{eq:case1}, we have
% $\frac{2}{c-1}L_{0\text{-}1}(\mathcal{P},G_{pl})+\frac{c+1}{c-1}R_{\alpha}(\mathcal{P})+2R_{\alpha}(\mathcal{N}) \leq \frac{2}{c-1}Err(G_{pl})+\frac{c+1}{c-1}\mu+2\omega=B_{\mathcal{P}}$. Then using $\omega\leq\frac{1}{2}\mu$ in \cref{ass:sep}, we obtain $$B_{\mathcal{P}}\leq \frac{2}{c-1}Err(G_{pl})+\frac{2c}{c-1}\mu=B_{\mathcal{F}}.$$
\begin{align}
Err(\widehat{\mathcal{P}})&\leq \frac{2}{c-1}Err(G_{pl})+\frac{c+1}{c-1}\mu+2\omega =B_{\mathcal{P}} \nonumber\\
      &\leq \frac{2}{c-1}Err(G_{pl})+\frac{2c}{c-1}\mu=B_{\mathcal{F}}. \tag{using $\omega\leq\frac{1}{2}\mu$ in \cref{ass:sep}}\nonumber
\end{align}
Likewise, when $L_{0\text{-}1}(\mathcal{P},G_{pl})+R_{\alpha}(\mathcal{P})> (c-1)Err(G_{pl})$, by \cref{eq:case2}, we have 
$$Err(\widehat{\mathcal{P}})\leq\frac{2}{c-1}Err(G_{pl})+\frac{2c}{c-1}\mu= B_{\mathcal{P}}=B_{\mathcal{F}}.$$
\end{proof}
Our results show that, by minimizing \cref{eq:mutex} to bound $Err(\mathcal{P})$, mutex-based consistency regularization (corresponding to  \cref{ass:set}) could help TPC improve the accuracy on pseudo-labels.

\begin{table*}[t]
   \caption{Accuracy (\%) on CIFAR-10/100 and SVHN averaged on 5 runs. 
   Results with $\mbox{}^\dagger$  are reported in FixMatch~\cite{sohn2020fixmatch} and results with $\mbox{}^\ddagger$ are reported in SLA~\cite{tai2021sinkhorn}, while results with $\mbox{}^\ast$ are using our own reimplementation.
%   and other results are reported in FixMatch~\cite{sohn2020fixmatch}. 
   Results of ReMixMatch and CoMatch are achieved by combining the \textit{distribution alignment (\textbf{DA})}~\cite{berthelot2020remixmatch}. We mark out the \textbf{best} and \colorbox{light-gray-2}{second best} accuracy. }
   \vskip 0in
   \resizebox{\textwidth}{!}{  
      \label{table1}
      \begin{threeparttable}
      \begin{tabular}{@{}lccccccccc@{}}     
      \toprule
      \multirow{2}{*}{Method} & \multicolumn{4}{c}{CIFAR-10}                                                                     & \multicolumn{3}{c}{CIFAR-100} & \multicolumn{2}{c}{SVHN} \\  \cmidrule(lr){2-5}  \cmidrule(lr){6-8}  \cmidrule(lr){9-10} 
                              & 10 labels                       & 20 labels                       & \multicolumn{1}{c}{40 labels} & \multicolumn{1}{c}{80 labels} & 200 labels & 400 labels    & 2500 labels   & 40 labels   & 250 labels \\ \cmidrule(r){1-1} \cmidrule(lr){2-5}  \cmidrule(lr){6-8}  \cmidrule(lr){9-10}  
      $\mbox{MixMatch \cite{berthelot2019mixmatch}}^\dagger$               & -                               & - & 52.46$\pm$11.50                   & -            & -         & 33.39$\pm$1.32    & 60.06$\pm$0.37    & 57.45$\pm$14.53  & 96.02$\pm$0.23 \\
      $\mbox{UDA \cite{xie2020unsupervised}}^\dagger$                     & -                               & -                               & 70.95$\pm$5.93                    & -                  & -   & 40.72$\pm$0.88    & 66.87$\pm$0.22    & 47.37$\pm$20.51  & 94.31$\pm$2.76  \\
      $\mbox{ReMixMatch \cite{berthelot2020remixmatch}}^\dagger$              & -                               & -                               & 80.90$\pm$9.64                    & -           & -          & 55.72$\pm$2.06    & \colorbox{light-gray-2}{72.57$\pm$0.31}    & \colorbox{light-gray-2}{96.66$\pm$0.20}  & 97.08$\pm$0.48  \\
    %   FixMatch w. DA          & -                 & $\mbox{83.81$\pm$9.35}^\dagger$  & $\mbox{86.98$\pm$3.40}^\dagger$                   & $\mbox{92.29$\pm$0.86}^\dagger$                      & -             & - & -             & -           & -          \\ 
    %   $\mbox{EnAET \cite{wang2020enaet}}^\ddagger$         & $\mbox{-}$  & $\mbox{-}$                   & \textbf{83.55}  & {90.65}                 & -             & -  & -              & 83.08           & 96.79         \\
      $\mbox{SLA \cite{tai2021sinkhorn}}^\ddagger$         & $\mbox{65.87$\pm$10.83}$  & $\mbox{81.91$\pm$6.77}$                   & \textbf{94.83$\pm$0.32}  & \colorbox{light-gray-2}{94.98$\pm$0.28 }                 & -             & \textbf{58.56$\pm$1.41}  & 71.27$\pm$0.44              & 96.37$\pm$2.91            & -          \\\cmidrule(r){1-1} \cmidrule(lr){2-5}  \cmidrule(lr){6-8}  \cmidrule(lr){9-10}  
      $\mbox{FixMatch \cite{sohn2020fixmatch}}^\ast$                & 45.91$\pm$28.46     & 84.97$\pm$10.37  & 89.18$\pm$1.54        & 91.99$\pm$0.71                & 38.87$\pm$2.50                   & 52.20$\pm$1.88    & 71.63$\pm$0.21    & 96.54$\pm$1.05  & 97.44$\pm$0.26  \\ 
      % AlphaMatch            & -                               & -                               & 91.35$\pm$3.38                    & 95.03$\pm$0.29                     & -             & -             & 97.03$\pm$0.26  & 97.56$\pm$0.32 \\
      $\mbox{CoMatch \cite{li2021comatch}}^\ast$              & \colorbox{light-gray-2}{69.87$\pm$11.82}                     & 88.43$\pm$7.22  & 93.21$\pm$1.55                    & 94.08$\pm$0.31         & 40.10$\pm$2.99            & \colorbox{light-gray-2}{58.04$\pm$1.39}             & 72.45$\pm$0.44              & 96.47$\pm$1.29           & 96.98$\pm$1.29           \\ \cmidrule(r){1-1} \cmidrule(lr){2-5}  \cmidrule(lr){6-8}  \cmidrule(lr){9-10}  
      MutexMatch              & \textbf{76.06$\pm$18.28}\tnote{1}         & \colorbox{light-gray-2}{91.77$\pm$2.60}            & 93.22$\pm$2.52           & 93.23$\pm$0.81       & \colorbox{light-gray-2}{40.38$\pm$2.36}     & 56.14$\pm$1.46            & 71.80$\pm$0.23            & \textbf{97.19$\pm$0.26}           & \textbf{97.73$\pm$0.18}          \\
      MutexMatch ($k=0.6C$)             & 57.52$\pm$21.17            & \textbf{92.23$\pm$3.23}             & \colorbox{light-gray-2}{94.21$\pm$0.84}           & \textbf{95.00$\pm$0.34}       & \textbf{41.59$\pm$1.86}     & 55.59$\pm$0.42             & \textbf{72.82$\pm$0.40}            & 96.55$\pm$1.46           & \colorbox{light-gray-2}{97.47$\pm$0.17}          \\ \bottomrule
      \end{tabular}
      \begin{tablenotes}
        \footnotesize
        \item[1] This result is achieved by DA. Notice that CoMatch also integrates DA technique into training, and we find that DA will greatly improve performance when the amount of labels is very small (\eg, 10 labels). Default MutexMatch achieves 66.45$\pm$30.42\% accuracy. 
        % More details about barely supervised learning can be found in \cref{ap_bare}. 
      \end{tablenotes}
   \end{threeparttable}
   }
    \vskip 0in
\end{table*}

\section{Experiments}
\label{exp}

In this section, we evaluate MutexMatch on five benchmark datasets, including  CIFAR-10/100 \cite{krizhevsky2009learning}, SVHN \cite{netzer2011reading}, STL-10 \cite{coates2011an}, mini-ImageNet \cite{vinyals2016matching} and Tiny-ImageNet.
We also conduct ablation studies  in \cref{ab_section} to investigate the efficacy of MutexMatch. Other experiments, \eg, the impact of consistency regularization on TNC, are shown  in \cref{sec:fd}.

\begin{table}[t]
    \centering
    \scriptsize
          \caption{ 
          Accuracy (\%) on CIFAR-10 and CIFAR-100 with larger amounts of labels. 
        %   Experiments are conducted on CNN-13 \cite{oliver2018realistic}  and Wide ResNet. 
       Results with $^{\dagger}$ are reported in UPS~\cite{rizve2021in} and results with $^{\ddagger}$ are reported in EnAET~\cite{wang2020enaet} whereas results with $^{\ast}$ are based on our reimplementation. Notably, for CIFAR-100, EnAET adopts a larger backbone: WRN-28-2 with 135 filters per layer (26M parameters) than WRN-28-8 (23M parameters) we used.}
          
            \label{cnn_table}
            \vskip 0in
        \setlength{\tabcolsep}{2.1mm}{
         \!\!\!\!\begin{tabular}{@{}lcccc@{}}      
          \toprule
          \multirow{2}{*}{Method} & \multicolumn{2}{c}{CIFAR-10} & \multicolumn{2}{c}{CIFAR-100}   \\ \cmidrule(lr){2-3} \cmidrule(lr){4-5} 
                                  & {1000 labels} & {4000 labels} & {4000 labels} & {10000 labels}\\ \cmidrule(r){1-1} \cmidrule(lr){2-3} \cmidrule(lr){4-5} 
          \multicolumn{5}{c}{\textbf{Backbone: CNN-13}} \\              
          $\mbox{MT \cite{tarvainen2017mean}}^{\dagger}$            & {80.96$\pm$0.51} & {88.59$\pm$0.25}& {54.64$\pm$0.49}& {63.92$\pm$0.51} \\
          $\mbox{ICT \cite{verma2019interpolation}}^{\dagger}$                 & {84.52$\pm$0.78}  & {92.71$\pm$0.02}&-&-\\
          $\mbox{DualStudent \cite{ke2019dual} }^{\dagger}$          & {85.83$\pm$0.38}  & {91.11$\pm$0.09}&-& {67.23$\pm$0.24}\\
          $\mbox{UPS \cite{rizve2021in} }^{\dagger}$                  & {91.82$\pm$0.15}  & {93.61$\pm$0.02}& {59.23$\pm$0.10}& {68.00$\pm$0.49}\\ \cmidrule(r){1-1} \cmidrule(lr){2-3} \cmidrule(lr){4-5} 
          MutexMatch              & \textbf{93.01$\pm$0.32}  & \textbf{94.10$\pm$0.24}& \textbf{63.09$\pm$0.52}& \textbf{68.52$\pm$0.31}\\ \midrule
          \multicolumn{5}{c}{\textbf{Backbone: Wide ResNet}} \\           
          $\mbox{EnAET \cite{wang2020enaet} }^\ddagger$               & {93.05} & {94.65}& {-}& {77.08} \\
          $\mbox{FixMatch \cite{sohn2020fixmatch}}^\ast$             & {94.86$\pm$0.10} & {95.56$\pm$0.09}& {74.07$\pm$0.19}& {77.48$\pm$0.18} \\
          $\mbox{CoMatch \cite{li2021comatch}}^\ast$                & {95.02$\pm$0.25} & \textbf{95.84$\pm$0.04}& {75.08$\pm$0.22}& {78.01$\pm$0.27} \\\cmidrule(r){1-1} \cmidrule(lr){2-3} \cmidrule(lr){4-5} 
          MutexMatch              & \textbf{95.35$\pm$0.33}  & {95.63$\pm$0.06}& \textbf{75.32$\pm$0.35}& \textbf{78.06$\pm$0.11}\\ \bottomrule
          \end{tabular}}
          
          \vskip 0in
    \end{table}

\begin{table}[t]
    \centering
    \scriptsize
            \caption{Accuracy (\%) on STL-10, mini-ImageNet and Tiny-ImageNet. 
            Results of $\mbox{}^\ast$ are reported in CoMatch~\cite{li2021comatch} and results of other baselines are reported on our reimplementation.}
            \label{table2}
            % \vskip 0.015in
            \setlength{\tabcolsep}{1.9mm}{
            \begin{tabular}{@{}lcccc@{}}      
            \toprule
            \multirow{2}{*}{Method} & \multicolumn{2}{c}{STL-10}  & \multicolumn{1}{c}{mini-ImageNet} & \multicolumn{1}{c}{Tiny-ImageNet}\\ \cmidrule(lr){2-3} \cmidrule(lr){4-4}  \cmidrule(lr){5-5} 
                        & {1000 labels} & {5000 labels} & {1000 labels} & {5000 labels}   \\ \cmidrule(r){1-1} \cmidrule(lr){2-3} \cmidrule(lr){4-4}  \cmidrule(lr){5-5}
            \multicolumn{5}{c}{\textbf{Backbone: ResNet-18}}\\
            \mbox{FixMatch \cite{sohn2020fixmatch}}& $\mbox{65.38$\pm$0.42}^\ast$ & {88.29$\pm$1.00} &39.03$\pm$0.66 &19.64$\pm$0.26\\
            \mbox{CoMatch \cite{li2021comatch}}  & $\mbox{79.80$\pm$0.38}^\ast$ &{90.56$\pm$0.22} &43.72$\pm$0.58 & 20.37$\pm$0.30 \\ \cmidrule(r){1-1}  \cmidrule(lr){2-3} \cmidrule(lr){4-4}  \cmidrule(lr){5-5} 
            MutexMatch     & \textbf{83.36$\pm$0.22}   &\textbf{91.15$\pm$0.17} &\textbf{48.04$\pm$0.52} &\textbf{23.55$\pm$0.21}\\ \midrule
            \multicolumn{5}{c}{\textbf{Backbone: Wide ResNet}} \\
            \mbox{FixMatch} & {92.46$\pm$0.35} & 95.54$\pm$0.18 & {58.45$\pm$0.20} &{42.05$\pm$0.34} \\
            \mbox{CoMatch }  & {91.50$\pm$0.53} & 95.00$\pm$0.30 & {59.88$\pm$0.39} & {43.26$\pm$0.45}\\\cmidrule(r){1-1}  \cmidrule(lr){2-3} \cmidrule(lr){4-4}  \cmidrule(lr){5-5} 
            MutexMatch                & \textbf{92.52$\pm$0.86} &\textbf{96.10$\pm$0.13} &\textbf{60.86$\pm$0.37} &\textbf{44.18$\pm$0.23}\\\bottomrule
            \end{tabular}
            }
            \vskip 0in
    \end{table}

\subsection{CIFAR-10, CIFAR-100 and SVHN}
\label{ci_section}
Firstly, we evaluate our method and baselines on three widely used SSL datasets: 
% CIFAR-10/100 and SVHN.
% (1) CIFAR-10, consisting of 10 classes, 50,000 images for training and 10,000 images for evaluation;
(1) CIFAR-10/100, consisting of 10/100 classes, 50,000 images for training, 10,000 images for evaluation, and
(2) SVHN, consisting of more than 70,000 street view house number images from 10 classes. 

\noindent\textbf{Baselines.} We introduce recent state-of-the-art SSL methods, \ie, SLA~\cite{tai2021sinkhorn}, CoMatch~\cite{li2021comatch}, FixMatch~\cite{sohn2020fixmatch} and FixMatch with distribution
alignment~\cite{berthelot2020remixmatch} to compare with MutexMatch. Moreover, we compare  with recently-proposed SSL methods such as UDA~\cite{xie2020unsupervised}, MixMatch~\cite{berthelot2019mixmatch} and ReMixMatch~\cite{berthelot2020remixmatch}. 

\noindent\textbf{Settings.}\quad For all experiments in MutexMatch, Wide ResNet~\cite{zagoruyko2016wide} is adopted as the backbone (WRN-28-2 for CIFAR-10 and SVHN,  WRN-28-8 for CIFAR-100) following~\cite{sohn2020fixmatch}. In our implementation, TNC is the same two-layer MLP as TPC. For fair comparison, We follow these baseline methods~\cite{sohn2020fixmatch,li2021comatch} using SGD with a momentum of 0.9 and a weight decay of 0.0005 during training. Also, we train the model for 1024 epochs, using a learning rate of 0.03 with cosine decay schedule. 
For $k$ in MutexMatch, two versions of the experiment are provided for comparison. One  we simply set $k$ as the total amount of classes $C$ (\ie, $k=10$ for CIFAR-10/SVHN, $k=100$ for CIFAR-100), which is denoted as \textit{MutexMatch} (this works as the setting for all the following experiments, unless noted otherwise); The other we set $k=6$ for CIFAR-10 and SVHN, $k=60$ for CIFAR-100, which is denoted as \textit{MutexMatch ($k=0.6C$)}.
For other hyper-parameters in MutexMatch,
we set $\tau =0.95, \mu =7, B=64$ for all experiments.
In our method, RandAugment~\cite{cubuk2020randaugment} is used for strong augmentation. Moreover, $\lambda_{sep}$, $\lambda_{p}$ and $\lambda_{n}$ are set to 1 for simplicity. Lately, to reduce the influence of random data partition, we report the mean
and standard deviation of accuracy on five different folds of labeled/unlabeled data.

\noindent\textbf{Results.} As shown in \cref{table1}, our approach obtains the highest accuracy under most settings.
With only 4 labeled data per class, our method achieves an accuracy of 94.21$\pm$0.84\% on CIFAR-10, and 97.19$\pm$0.26\% on SVHN, yielding improvement over prior SSL results. Especially, we show the superiority of MutexMatch under the extremely label-scarce setting.
\eg, achieving an average accuracy of 92.23\% on CIFAR-10 with 20 labels, 41.59\% on CIFAR-100 with 200 labels. 
Usually fewer labels will result in the accumulation of more noisy pseudo-labels in training, yet, MutexMatch is able to utilize all unlabeled data while introducing little noise. In another word, our approach helps the model to be on track as early in the training as possible due to the benefit of a more reliable information gain, \ie, consistent regularization on complementary labels with high accuracy, which leads to a robust performance in the end.

Further, the comparisons with more baseline methods are reported on CIFAR-10/100 with more backbones (CNN-13 \cite{oliver2018realistic} is provided) and more available labels. 
% We further compare MutexMatch with more baselines.
% MT~\cite{tarvainen2017mean}, ICT~\cite{verma2019interpolation}, DualStudent~\cite{ke2019dual} and UPS~\cite{rizve2021in}. 
The experimental settings are the same as that of CIFAR-10/100 mentioned above except that  $\tau=0.5$.
\cref{cnn_table} shows MutexMatch is backbone independent, 
% and it outperforms all baselines.
and it achieves performance improvement when more labels are given, outperforming most baselines. In \cref{cnn_table}, the comparisons with UPS \cite{rizve2021in}, a recently proposed pseudo-labeling based SSL method, are worth focusing on, because UPS also implicitly benefits from low-confidence samples, \ie, introducing negative cross-entropy
loss (NCE) to learn the information of complementary label. Holding the same experimental setup as UPS, we show MutexMatch's exploitation of low-confidence samples has advantages over UPS. More discussions on this exploitation can be found in \cref{sec:uls}.

\subsection{STL-10,  mini-ImageNet and Tiny-ImageNet}
 \vskip 0.03in
STL-10 contains 5,000 labeled from 10 classes and 100,000 unlabeled images extracted from a similar but broader distribution.  mini-ImageNet and Tiny-ImageNet are subsets of ImageNet \cite{deng2009imagenet}, which contain 60,000/100,000 images evenly distributed across 100/200 classes respectively.
The challenge of the three datasets lies in the fact that unlabeled images contains out-of-distribution images or with a larger number of categories, which enables us to test the robustness of MutexMatch.

\noindent\textbf{Settings.} For STL-10, MutexMatch is evaluated on the 5 folds predefined in original dataset, with each fold containing 1,000 labeled data, and the averaged results are reported. Meanwhile, the results using total 5 folds (\ie, 5000 training samples) as labeled data are also reported.
% The final result is achieved by training five models and averaging their performances. 
For mini-ImageNet and Tiny-ImageNet, we report the results with 1000 labels and 5000 labels averaged on five folds, respectively.
As same as \cref{ci_section}, firstly, we adopt Wide ResNet (WRN) for the three datasets, \ie, WRN-37-2 for STL-10, WRN-28-8 for mini-ImageNet and Tiny-ImageNet. Additionally, following~\cite{li2021comatch}, we report results with a lighter backbone network: ResNet-18 \cite{he2016deep} to provide more comprehensive comparisons with baseline methods.
% as the backbone because it consumes less computation cost. 
We use the same hyper-parameters and learning rate as  in \cref{ci_section}, and train the models using SGD with a momentum of 0.9 and a weight decay of 0.0005.

\noindent\textbf{Results.} As shown in \cref{table2}, MutexMatch's performance advantage is maintained across all backbones.
For example, with 1000 labels and ResNet-18 backbone, compared with CoMatch, MutexMatch achieves accuracy improvement from 79.80$\pm$0.38\% to 83.36$\pm$0.22\% on STL-10 and from 43.72$\pm$0.58\% to 48.04$\pm$0.52\% on mini-ImageNet. 
% This means the performance of MutexMatch is far superior than that of the existing methods and the reasons for this can be summarized as: 
From \cref{table2}, we find our approach is more prominent in two scenarios:
1) When lighter models are employed (\eg, ResNet-18 in \cref{table2}), the models themselves have less capacity and are less able to learn from unlabeled data. And using large models is already able to learn enough information from high-confidence samples. Therefore, when using small models, the additional information that can be provided by learning low-confidence samples using complementary labels is more substantial and can help improve the performance of the model more effectively; 2) When the dataset is harder (\eg, mini-ImageNet and Tiny-ImageNet in \cref{table2}), it is more difficult for the model to confidently predict the pseudo-label. After training for a considerable period of time, there are still most of the samples possessing low confidence levels. This means that the approaches of using a predefined threshold to filter pseudo-labels will waste a large number of unlabeled samples and eventually lead to poor performance. In contrast, MutexMatch does not spare a single unlabeled data and allows the model to learn as much information as possible.

\section{Ablation Study}
\newcommand{\mz}{5.5cm}
\label{ab_section}
Extensive ablation studies have been conducted to verify the effectiveness of MutexMatch. 
In the experiments mainly conducted on CIFAR-10 and SVHN using 4 labels per class, MutexMatch achieves 93.22$\pm$2.52\% and 97.19$\pm$0.26\% accuracy respectively using default setting. In the following experiments (as well as in \cref{sec:fd}), we use the same settings as in \cref{ci_section} ($k=C$) and average the results over multiple runs, where we keep the supervised loss as Equation \eqref{s} and the positive consistency loss as Equation \eqref{p}.
More additional results are available in \cref{sec:fd}.

\subsection{Utilization of Low-confidence Samples}
\label{sec:uls}
\begin{figure}[t]
% % \setlength{\abovecaptionskip}{-0.1em}
  \centering
%   \hspace{-3mm}
  \subfloat[]{
   \includegraphics[width=4cm,height=3.5cm]{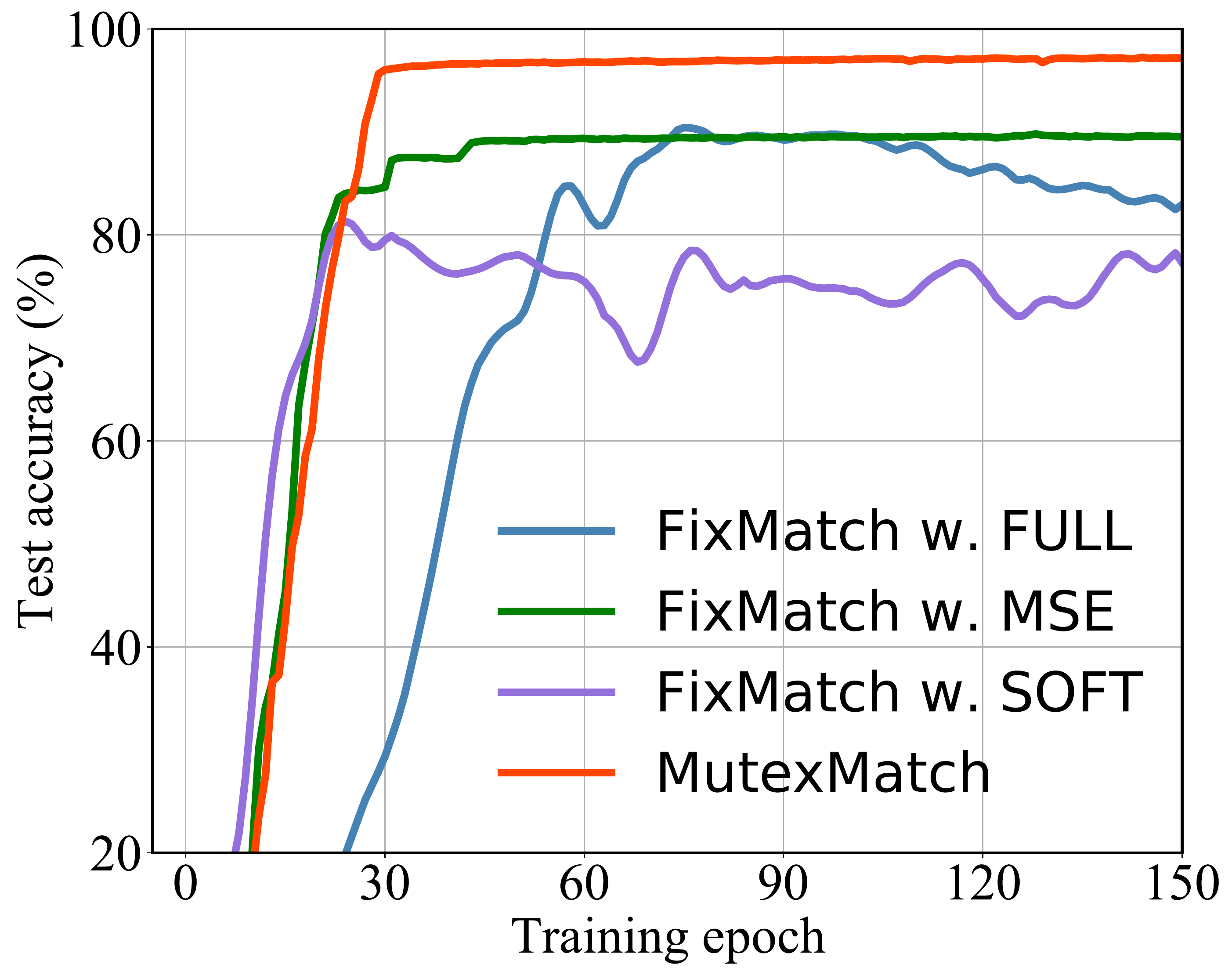}
   \label{fig:short-a}
   }
  % \hspace{-2mm}
% \hfil
   \subfloat[]{
   \includegraphics[width=4cm,height=3.5cm]{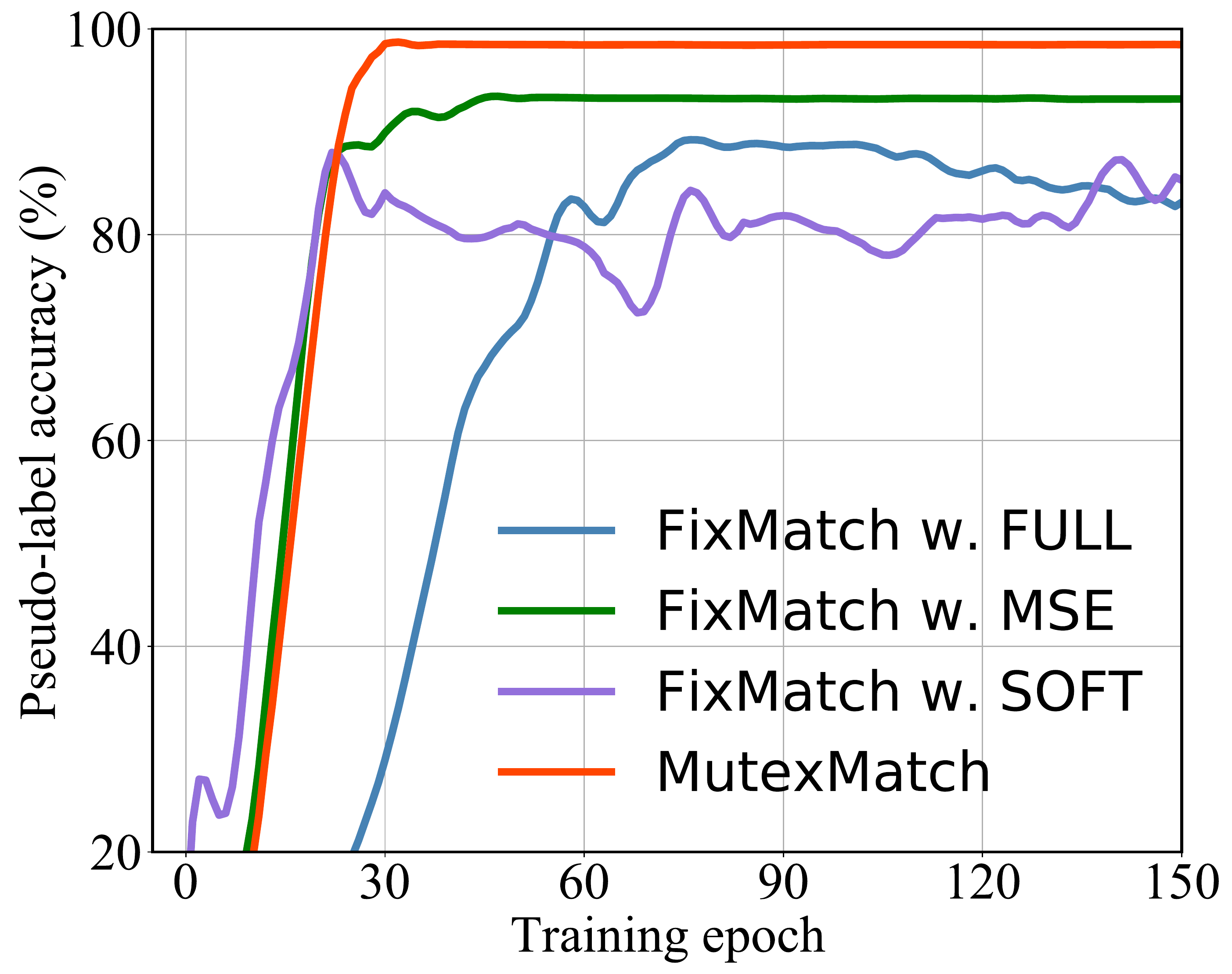}
   \label{fig:short-b}
   }
% \vspace{-1em}
  \caption{Ablation study on SVHN with 40 labels. The x-axis represents the training epoch , whereas y-axis in (a) represents the test accuracy and in (b) the accuracy of pseudo-labels.}
  \label{fig2}
  \vskip 0in
\end{figure}

We believe that the reason MutexMatch outperforms other earlier SSL algorithms is because TNC enables the model to learn from all unlabeled data. 
For example, in FixMatch, due to a predefined confidence threshold, the unlabeled samples with smaller confidence than this threshold will not participate in the training. 
Therefore, we use the three most intuitive ways to use all the unlabeled data for ablation study.
We first use TPC to compute prediction $p^{w}=\hat{\mathcal{P}} (\hat{\theta}(x^{w}))$ of weakly-augmented unlabeled data $x^{w}$ and then:
\begin{enumerate}
  \item [(\romannumeral1)] 
  We use $\hat{p}^{w}=\arg \max (p^{w}) $ as a hard pseudo-label and enforce the cross-entropy loss against the model’s prediction $p^{s}=\mathcal{P} (\theta(x^{s}))$ of $x^{s}$: 
\begin{equation}\mathcal{L}_{ab1}=\frac{1}{\mu B}\sum_{n = 1}^{\mu B}\mathbbm{1}(\max (p^{w}_{n})<  \tau )H(\hat{p}^{w}_{n},p^{s}_{n}).   \end{equation}       
  \item [(\romannumeral2)]
  We use $p^{w}$ as a soft pseudo-label and enforce the cross-entropy loss against $p^{s}$:
\begin{equation}\mathcal{L}_{ab1}=\frac{1}{\mu B}\sum_{n = 1}^{\mu B}\mathbbm{1}(\max (p^{w}_{n})<  \tau )H(p^{w}_{n},p^{s}_{n}).   \end{equation}
  \item [(\romannumeral3)]
  We first obtain the feature $z^{w}=\theta (x^{w})$ of $x^{w}$ and the feature $z^{s}=\theta (x^{s})$ of $x^{s}$ extracted by the feature extractor $\theta$, and then we compute:
\begin{equation}\mathcal{L}_{ab1}=\frac{1}{\mu B}\sum_{n = 1}^{\mu B}\mathbbm{1}(\max (p^{w}_{n})<  \tau )E(z^{w}_{n},z^{s}_{n}),   \end{equation}
where $E(p,q)$ denotes the mean squared loss between the two distributions $p$ and $q$.
\end{enumerate}
The above loss  minimized by experiments is simply $\mathcal{L}_{sup}+\mathcal{L}_{p}+\mathcal{L}_{ab1}$. 
All models are trained on SVHN using four labels per class, and  the results of all experiments are shown in \cref{fig2}, where \textit{FULL} indicates the setting of (\romannumeral1), \textit{SOFT} indicates the setting of (\romannumeral2) and \textit{MSE} indicates the setting of (\romannumeral3), respectively.
On this dataset, the default MutexMatch achieves an accuracy of 97.19$\pm$0.26\%, outperforming experiments with other settings. Other ways using low-confidence samples could introduce more noisy pseudo-labels, which is unconducive to the consistency regularization. 
In addition, we conduct further experiments on SVHN for the setting of (\romannumeral1) where a weight $\lambda_{ab1}$ is used to adjust the importance of $\mathcal{L}_{ab1}$. As show in \cref{tab:r1}, despite varying $\lambda_{ab1}$, using regular consistency loss to learn low-confidence samples also impairs the performance of original FixMatch (\ie, $\lambda_{ab1}=0$).

 \newcommand{\mysize}{5.4cm}
\begin{table}[t]
%   \vspace{-1em}
   \centering
%   \scriptsize
   \caption{Accuracy (\%) on SVHN with 40 labels and various $\lambda_{ab1}$ under the setting of (\romannumeral1). $\lambda_{c}$ represents the loss weight of $\mathcal{L}_{ab1}$ while $\mathcal{L}_{ab1}=0$ means original FixMatch.}
   \label{tab:r1}
%   \vskip 0.1in
 \setlength{\tabcolsep}{2.3mm}{

   \begin{tabular}{@{}lcccccc|c@{}}
   \toprule
   $\lambda_{ab1}$  & 0  & 0.1      & 0.2     & 0.5     & 1  & 2  & Ours  \\ \midrule
   Accuracy&96.54 & 92.89 & 89.26 & 89.16 & 88.40 &87.57  & 97.19 \\\bottomrule
   \end{tabular}}
\end{table}
% \subsection{Utilization of Low-confidence Samples}
% \label{sec:uls}
Meanwhile, we notice that UPS \cite{rizve2021in} also utilizes the complementary labels to exploit low-confidence samples. \cite{rizve2021in} introduces negative cross-entropy loss (NCE) to learn the information of complementary labels. We use TNC to decouple the part where complementary labels are directly involved in training. 
And then we use soft labels to enforce consistency regularization on TNC so as TNC could help the model learn a better data representation of unlabeled data. For fair comparison, We first conduct the following experiments on CIFAR-10 with 40 labels, which is built on the paradigm of FixMatch. Then we provide a further discussion on the reason for not using NCE to learn low-confidence samples.
\begin{enumerate}
  \item [(\romannumeral4)] 
  We remove TNC and select the positive label and the complementary label as in \cite{rizve2021in}. We introduce NCE to learn the information of complementary label (also in the same way as in \cite{rizve2021in}). This is equivalent to the following setting: keeping TPC, for the image $x$, TPC outputs prediction $p^{w}$ on the weakly-augmented version of $x$. Then we select the positive label and the complementary label through the threshold $\tau_{p}$ and $\tau_{n}$ \cite{rizve2021in}. We enforce consistency regularization between the positive label and prediction $p^{s}$ on strongly-augmented version of $x$, and then enforce NCE between the complementary label and $p^{w}$ at the same time. We set the same $\tau_{n}$ as \cite{rizve2021in}, then we set $\tau_{p}=0.95$ (\ie, $\tau$ in MutexMatch), $\tau_{p}=0.7$ (\ie, $\tau_{p}$ in \cite{rizve2021in}) for two experiments, which are denoted as \textit{F w. NCE (1)} and \textit{F w. NCE (2)} respectively.      
  \item [(\romannumeral5)]
  Instead of using NCE to learn complementary label directly, we enforce consistency  between $p^{w}$ and $p^{s}$ by NCE (denoted as \textit{F w. NCE (3)}).
\end{enumerate}

As shown  in \cref{tb:nce}, we can see MutexMatch outperforms other settings utilizing low-confidence samples with NCE. We analyze that the direct use of NCE to learn complementary label like \cite{rizve2021in} will lead to the ``homogenization'' of the learned information, \ie, the cross-entropy loss on the positive label and the NCE loss on the complementary labels share predictions of the same classifier. Moreover, the probability of the element (corresponding to complementary labels selected by the threshold ) in prediction is very close to $0$. The threshold $\tau_{n}$ is small because \cite{rizve2021in} needs to ensure the accuracy of the complementary label. In this case, the impact of the loss item of NCE is very small. This way of using low-confidence samples may not be informative, hence not helpful for the model, which means that the final performance is still at the FixMatch level (even lower), as can be confirmed by the results of \textit{F w. NCE (1)$\sim$(3)} shown  in \cref{tb:nce}. Even if \cite{rizve2021in} uses a very small threshold $\tau_{n}$, the accuracy of the complementary labels selected in the same way as in \cite{rizve2021in} is lower than that of the complementary labels output by TNC in MutexMatch. Considering that \cite{rizve2021in} treats complementary label as hard label and we treat that as soft label, for fair comparison, we provide the results of \textit{MutexMatch (Hard-Hard)}, which is described in \cref{abl:strategy} (hard complementary labels are used for separate training of TNC and consistency regularization on TNC).

 \begin{table}[t]
%   \vspace{-1em}
   \centering
   \footnotesize

   \caption{Ablation study on CIFAR-10 with 40 labels, which is corresponding to the settings of (\romannumeral4) and (\romannumeral5) in \cref{sec:uls}. CPA indicates complementary pseudo-label accuracy. } 
   \vskip 0in
   \label{tb:nce}
   \setlength{\tabcolsep}{2.6mm}{
\begin{tabular}{@{}ccccc@{}}
\toprule
Ablation               & $\tau_{p}$ & $\tau_{n}$ &CPA (\%) & Accuracy (\%) \\ \midrule
F w. NCE (1)           & 0.95       & 0.05       & 98.84                               & 87.82    \\
F w. NCE (2)           & 0.7        & 0.05       & 97.73                               & 78.14    \\
F w. NCE (3)           & 0.95       & 0.05       & 98.62                               & 84.33    \\ \midrule
MutexMatch (Hard-Hard) & 0.95       & -          &\textbf{99.96}                               & 90.56    \\
MutexMatch             & 0.95       & -          & -                              & \textbf{93.22}    \\ \bottomrule
\end{tabular}
}
\vskip 0in
   \end{table}

\subsection{Ablation on TPC and TNC}
\label{abl:tt}
Since TPC is not well trained in the early stage of learning due to excessive noisy pseudo-labels, directly introducing complementary labels in this process is not helpful and even harmful. 
% the risk of eventually leading to training collapse.
Therefore, we use consistency regularization on TNC to \textit{``decouple''}  the part where complementary labels are directly involved in training, and stop gradient on $\theta$ during the separate training of TNC described in \cref{tnc_section}.  

For further discussion on the effectiveness of TPC and TNC, we consider removing TPC, TNC and the stop-gradient setting respectively.  As mentioned above, our design of the following experiments is shown in \cref{tb:ab}:  
(\romannumeral1) explores what happens when we train TNC with complementary labels generated by TPC without stopping the gradient on $\theta$; (\romannumeral2) indicates the results when we abandon the consistency on TNC. Restoring the back-propagation on  $\theta$ in Equation \eqref{sep} is a more reasonable setting, because if we set $\lambda_{n}=0$, then TNC will not participate in training at all, \ie, the setting of (\romannumeral3); (\romannumeral4) represents that we abandon the consistency on TPC; and  (\romannumeral5)$\sim$(\romannumeral8) together present further ablation studies of each component in MutexMatch. 

As shown  in \cref{tb:ab}, the default MutexMatch achieves overwhelmingly superior performance compared with other settings. 
% In figures, \textit{wo. TNC} represents setting of (\romannumeral1), \textit{wo. SG} represents setting of (\romannumeral2), \textit{wo. TPC} represents setting of (\romannumeral3) and \textit{w. 000}, \textit{w. 010}, \textit{w. 110}  represent setting of (\romannumeral4). 
Apparently, if TNC participates in model training directly, it will cause the collapse of training, which means that the model will be severely affected by the learning of TNC, so there is no way to learn effective information of ``what it is''. Both (\romannumeral1) and (\romannumeral2) show the superiority of using consistency regularization on TNC for learning of complementary label. This \textit{``decoupling''} ensures that TNC allows the model to learn a better data representation without adversely affecting the learning of TPC. (\romannumeral3) shows that TPC in this case fails to both get adequate training and complete the classification task, \ie, the training of TPC is similar to that using only labeled data (only by $\mathcal{L}_{sup}$). Meanwhile, the training of TNC is closely related to that of TPC. In this case, TNC has not been well trained either.  
Finally, the combination of (\romannumeral4)$\sim$(\romannumeral6) together proves the necessity of each component in MutexMatch. 

  \begin{table}[t]
%   \vspace{-1em}
   \centering
  \footnotesize
   \setlength{\belowcaptionskip}{1em}
   \caption{Ablation study on SVHN with 40 labels. 
%   \textit{SG} means stop gradient, and 
\XSolidBrush~in the fifth column indicates that we restore the gradient back-propagation on feature extractor $\theta$ in Equation \eqref{sep}. Notably, the settings of (\romannumeral3) and (\romannumeral7) are equivalent to original FixMatch.
%   More ablation studies on $\lambda_{p}$, $\lambda_{n}$ and $\lambda_{sep}$ can be found in \cref{sec:hp}.
} 
   \vskip 0in
   \label{tb:ab}
   \setlength{\tabcolsep}{2.2mm}{
\begin{tabular}{@{}cccccc@{}}
\toprule
Ablation            & $\lambda_{p}$ & $\lambda_{n}$ & $\lambda_{sep}$ & Stop gradient on $\theta$&Accuracy (\%)   \\ \midrule
(\romannumeral1)   & 1             & 1             & 1               & \XSolidBrush & 15.84\\
(\romannumeral2)   & 1             & 0             & 1               & \XSolidBrush & 14.79\\
(\romannumeral3)    & 1             & 0             & 1               & \CheckmarkBold& 96.54  \\
(\romannumeral4)   & 0             & 1             & 1               & \CheckmarkBold& 17.71 \\
(\romannumeral5)    & 1             & 1             & 0               & \CheckmarkBold& 83.24   \\
(\romannumeral6)    & 0             & 1             & 0               & \CheckmarkBold& 16.23  \\
(\romannumeral7)    & 1             & 0             & 0               & \CheckmarkBold& 96.54  \\
(\romannumeral8)   & 0             & 0             & 0               & \CheckmarkBold& 15.55   \\ \midrule
MutexMatch   & 1             & 1             & 1               & \CheckmarkBold& 97.19   \\ \bottomrule
\end{tabular}}

\vskip 0in
   \end{table}
 \begin{figure}[t]
 \vskip 0in

  \centering
  \subfloat[]{
   \includegraphics[width=3.9cm,height=3.44cm]{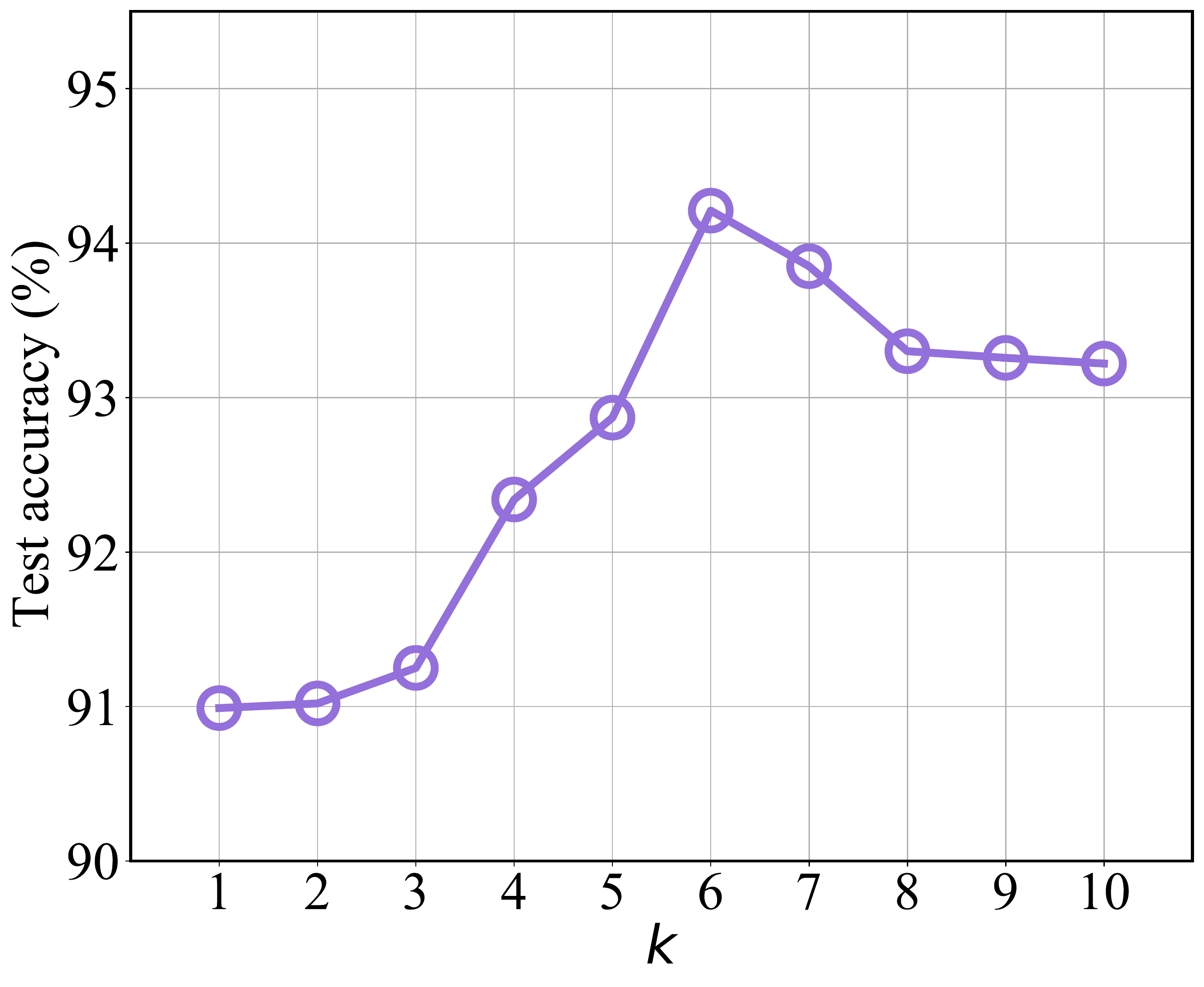}
   \label{fig:ab_k} 
   }
%   \hspace{-3mm}
\hfil
   \subfloat[]{
   \includegraphics[width=3.9cm,height=3.5cm]{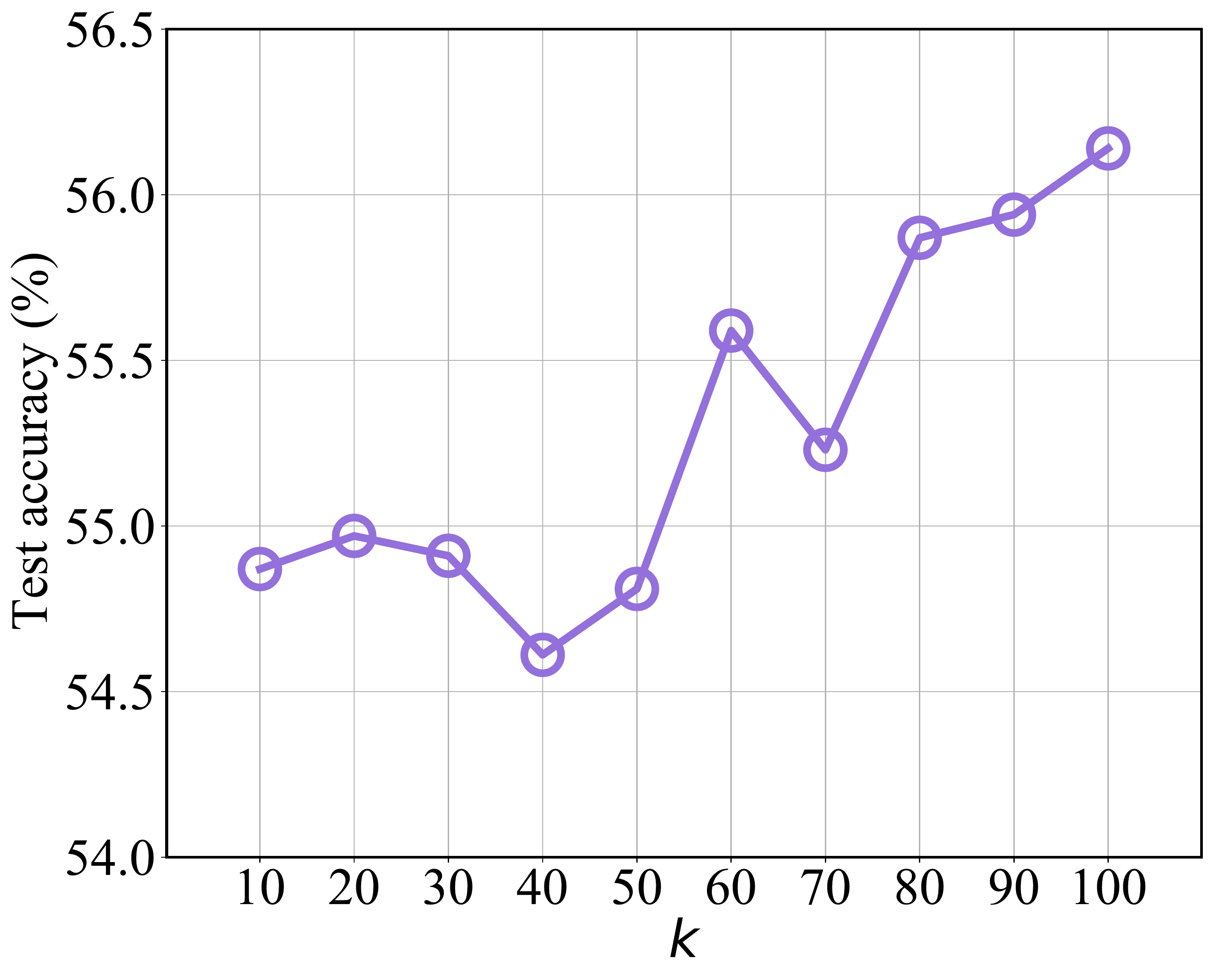}
   \label{fig:ab_k2}
   }
%   \vspace{-1em}
  \caption{Ablation study on $k$ controlling the consistency intensity on TNC.  (a) shows the results on CIFAR-10 with 40 labels, whereas (b) shows the results on CIFAR-100 with 400 labels. The results are averaged on multiple runs.}
  \label{fig:abk}
  \vskip 0in
\end{figure}

\subsection{Consistency Regularization on TNC}
\label{subsec:cr}
Firstly, we conduct ablation studies on hyper-parameter $k$ used for the intensity control of consistency on TNC. As the results in \cref{fig:abk} indicates, an appropriate $k$ will be more conducive to learning from low-confidence samples. Excessive consistency regularization may cause the model to overly narrow the overlap of the dissimilarity degree of low-confidence samples, consequently worsening the performance of the model. Whereas, too weak consistency regularization would provide too little guidance information for the model. We  note that MutexMatch is more sensitive to $k$ on CIFAR-10 than on CIFAR-100. This is easy to understand because CIFAR-10 is too simple and different settings are more likely to produce performance fluctuations on it. Actually, using the default MutexMatch (\ie, $k=C$) is sufficient to achieve good enough performance. Moreover, we conduct further ablation studies on consistency regularization on TNC as follows:
\begin{enumerate}
  \item [(\romannumeral1)] 
  We enforce consistency regularization on TNC over all samples, instead of enforcing only on low-confidence samples like the default setting.     
  \item [(\romannumeral2)]
  We use hard labels for the consistency regularization on TNC, \ie, we set the component $r^{w}_{n,(i)}=1$ in $r^{w}_{n}$ when $g_{n,(i)}=1$ for Equation \eqref{eq:ln}.
\end{enumerate}

As shown  in \cref{tb:crt}, the result of (\romannumeral1) confirms our claim that the consistency on TNC is more suitable for low-confidence samples. For high-confidence samples, it is sufficient to use consistency regularization on TPC to learn the guidance information provided by its pseudo-labels, while consistency regularization on complementary labels is superfluous. The result of (\romannumeral2)  proves that the consistency regularization on the TNC using hard labels is too strong, and since it does not contain the discriminative information of the degree of dissimilarity, it is not beneficial for the model performance. 
% To sum up, the consistency regularization on TNC is designed with the consideration  of learning complementary labels informatively and appropriately. 
More discussions of consistency regularization on TNC can be found in \cref{dis_section}. Additionally, more ablation studies on learning scheme of TNC can be found  in \cref{abl:strategy}.
 \begin{table}[t]
%   \vspace{-1em}
   \centering
  \footnotesize
    
   \caption{Ablation study on consistency regularization on TNC, which is corresponding to the settings of (\romannumeral1) and (\romannumeral2) in \cref{subsec:cr}. Results are reported on CIFAR-10 with 40 labels.} 
   \vskip 0in
   \label{tb:crt}
   \setlength{\tabcolsep}{2.1mm}{
\begin{tabular}{@{}cccccc@{}}
\toprule
Ablation               & $k$  & Pseudo-label accuracy (\%) & Accuracy (\%) \\ \midrule
MutexMatch  w. (\romannumeral1)           & 6   &92.13                             & 89.44    \\
MutexMatch  w. (\romannumeral2)           & 6   &85.97                           & 84.23    \\
MutexMatch         & 6                         &96.11     & 94.21    \\\midrule
MutexMatch w. (\romannumeral1)   & 10           &92.34                  & 90.35    \\
MutexMatch               & 10                   &95.09                 & 93.22    \\ \bottomrule
\end{tabular}
}
\vskip 0in
   \end{table}

\subsection{Hyper-parameters}
\label{sec:hp}

For MutexMatch, we should be very cautious about the choice of $\tau$, because different $\tau$ will lead to the division of high and low-confidence portions, which will in turn affect the impact of the mutex-based consistency regularization on the model. 
We vary $\tau$ to verify the sensitivity of MutexMatch to this hyperparameter.  
As shown  in \cref{tau}, MutexMatch needs to select appropriate $\tau$ to divide confidence portions. 
We note that $\tau$ has a greater impact on performance. The more labels are available, the less confirmation bias will be when using TPC directly for classification, 
so the portion of TPC in mutex-based consistency regularization can be used directly for learning. 
Therefore, we speculate that in general, we should choose a smaller $\tau$ to allow more pseudo-labels to participate in the training of TPC when the amount of labels increases.

\begin{table}[t]
   \centering
   \footnotesize
   \caption{Ablation study on confidence threshold $\tau$. Results are reported on CIFAR-10 varying amount of labels. } 
   \label{tau}
%   \vskip 0.1in
 \setlength{\tabcolsep}{1.9mm}{
   \begin{tabular}{@{}lcccc@{}}
   \toprule
   $\tau$                  & Labels        & Backbone             &Pseudo-label accuracy (\%)                & Accuracy (\%) \\ \midrule
   0.5                     & 10            & WRN-28-2             &85.35                & \textbf{79.83}    \\ 
   0.95                    & 10            & WRN-28-2             &83.99              & 66.45    \\\midrule
   0.5                     & 20            & WRN-28-2           &89.76                  & 88.59    \\ 
   0.95                    & 20            & WRN-28-2           &94.23                  & \textbf{91.77}    \\\midrule
   0.5                     & 40            & WRN-28-2           &90.50                  & 89.43    \\
   0.75                    & 40            & WRN-28-2           &91.99                  & 90.20    \\
   0.95                    & 40            & WRN-28-2           &95.09                        & \textbf{93.22}    \\ 
   0.99                    & 40            & WRN-28-2           &95.89                         & 92.17    \\ \midrule 
   0.5                     & 80            & WRN-28-2           &96.60                          & \textbf{93.80}    \\
   0.95                    & 80            & WRN-28-2           &96.55                          & 93.23    \\\midrule
   0.5                     & 1000          & CNN-13             &94.01                          & \textbf{93.01}    \\
   0.95                    & 1000          & CNN-13             &93.99                          & 92.02    \\\midrule
   0.5                     & 4000          & CNN-13             &95.32                          & \textbf{94.10}    \\
   0.95                    & 4000          & CNN-13             &94.64                          & 92.88    \\\bottomrule
   \end{tabular}
   }
%   \vskip -0.1in
\end{table}

\begin{table}[t]
%   \vspace{-1em}
  \centering
  \footnotesize

  \caption{Ablation studies on learning rate and learning rate schedule. Results are reported on CIFAR-10 varying amount of labels.} 
     \vskip 0.1in
  \label{ap_lr}
   
  \begin{tabular}{@{}lcccc@{}}
  \toprule
  Decay Schedule       & Learning Rate   & Labels   & Backbone                             & Accuracy (\%) \\ \midrule
  No Decay             & 0.03       & 40            & WRN-28-2                                    & \textbf{92.18}   \\
  No Decay             & 0.07       & 40            & WRN-28-2                                    & 92.01    \\ 
  No Decay             & 0.10       & 40            & WRN-28-2                                    & 91.66    \\ \midrule 
  Cosine Decay         & 0.03       & 40            & WRN-28-2                                    & \textbf{93.22}    \\
  Cosine Decay         & 0.07       & 40            & WRN-28-2                                    & 93.20    \\ 
  Cosine Decay         & 0.10       & 40            & WRN-28-2                                    & 92.59    \\ \midrule 
  No Decay             & 0.03        & 80            & WRN-28-2                                     & 93.03    \\
  Cosine Decay         & 0.03           & 80         & WRN-28-2                                     & \textbf{93.23}    \\\midrule
  No Decay             & 0.03        & 1000          & CNN-13                                       & 92.02    \\
  Cosine Decay         & 0.03           & 1000       & CNN-13                                       & \textbf{93.01}    \\\midrule
  No Decay             & 0.03        & 4000          & CNN-13                                       & 92.90    \\
  Cosine Decay         & 0.03           & 4000       & CNN-13                                       & \textbf{94.10}    \\\bottomrule
  \end{tabular}
% \vskip -0.1in
  \end{table}
\subsection{Learning Rate and Learning Rate Schedule}
We note that learning rate and learning rate schedule are very important for MutexMatch. 
In this section, we conduct additional ablation experiments for both. 
Following \cite{loshchilov2016sgdr}, recent works~\cite{sohn2020fixmatch,li2021comatch} use a cosine learning rate decay to achieve best performance. 
Likewise, as shown  in \cref{ap_lr}, we find that MutexMatch achieves better results with cosine learning rate decay on CIFAR-10. 
\eg, using 40 labels, MutexMatch training with cosine learning rate decay outperforms that training without learning rate decay by 1.04\%. 

\section{Further Discussions on TNC}
\label{sec:fd}

\subsection{Effectiveness Analysis of TNC}
\label{sec:atnc}
\begin{figure*}[t]
    % % \setlength{\abovecaptionskip}{-0.1em}
   \begin{center}
      \end{center}
    \centering
    \includegraphics[width=0.9\textwidth]{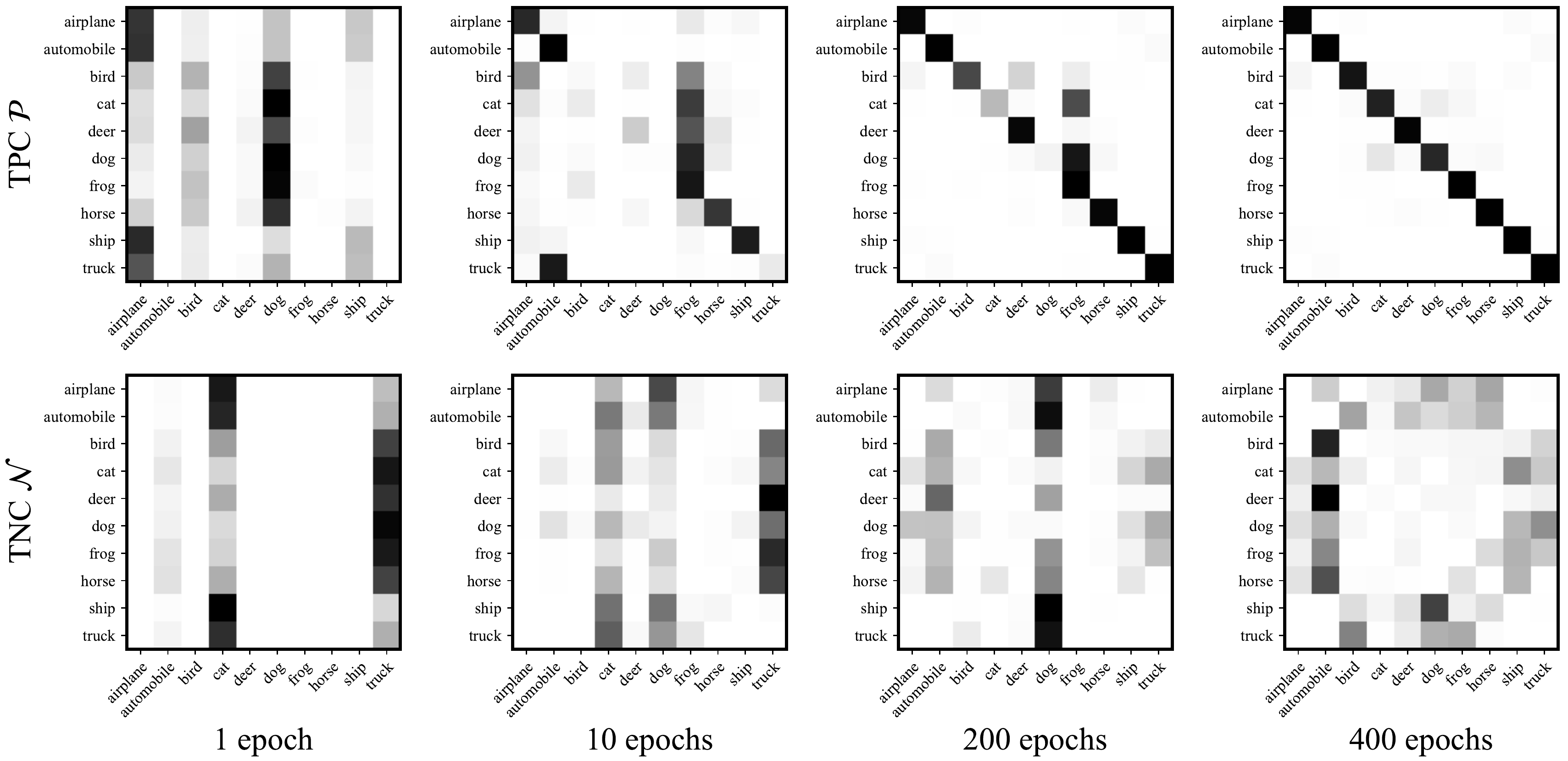}
    %   \vspace{-1em}
   \caption{The rate (\%) of each class (column in heat map) in the pseudo-labels and complementary pseudo-labels outputted by TPC and TNC respectively corresponding to each class (row in heat map) in CIFAR-10. The darker, the higher. 
   Results are reported on CIFAR-10 with 40 labels.} 
   \label{dis}
\vskip 0in
\end{figure*}

Compared with other baseline methods shown  in \cref{table1}, 
MutexMatch performs better on CIFAR-10 with extremely scarce labels. 
We believe that confirmation bias~\cite{yu2018learning} leads to the poor
performance of other methods. Fewer labels will introduce more noisy pseudo-labeled examples to participate in the learning process. 
Nevertheless, MutexMatch utilizes the unlabeled samples with low-confidence, in a more reasonable manner by TNC, introducing few noisy pseudo-labels.
As shown in \cref{ap1}, our experiments on CIFAR-10 show that MutexMatch produces more accurate pseudo-labels than FixMatch~\cite{sohn2020fixmatch}, especially when labeled samples are scarce.

Ideally, we think that for every class, unlike pseudo-labels from TPC, 
the distribution of complementary pseudo-labels from TNC should be evenly dispersed or diverse,
so MutexMatch can learn more multi-class information as much as possible. For example, considering CIFAR-100 (containing 100 classes), if the model only learns that the class
``monkey'' is not class ``truck'', then its revenue from the complementary labels outputted by TNC will not be very informative. 
As shown in \cref{dis}, we observe that the predictions from TNC are indeed generally consistent with our hypothesis. During the training, for each class of CIFAR-10,
the predictions of TNC are gradually dispersed to several classes (\ie, far away from the main diagonal of heat map), 
instead of gathering at a single class, indicating that TNC could output various complementary pseudo-labels for one class. 
On the contrary, the predictions outputted by TPC are gradually concentrated to the correct class (\ie, gathered to the main diagonal of the heat map). 
\begin{table}[t]
%   \vspace{-1em}
   \centering
   \footnotesize
   \caption{Accuracy (\%) of pseudo-label on CIFAR-10 with different amounts of labeled data.} 
   \label{ap1}
%   \vskip 0.1in
   \begin{tabular}{@{}lccccccc@{}}
   \toprule
   Labels    & 10      & 20     & 40     & 80  & 250  &1000   &4000   \\ \midrule
   FixMatch & 64.35 & 90.83 & 91.04  & 93.49 &96.35 &97.34  &97.44 \\ 
   MutexMatch & 83.99 & 94.23 & 95.09 & 96.55 &96.64 &97.33  &97.46\\\bottomrule
   \end{tabular}
% \vskip -0.1in
   \end{table}
\subsection{Ablation Study on the Learning Scheme of TNC}
\label{abl:strategy}
The learning of TNC is very important for MutexMatch. In default MutexMatch, we use hard complementary pseudo-label  $\hat{q}^{w}=\arg \min (p^{w}) $ to train TNC separately when stopping gradient back-propagation on the feature extractor, 
and enforce consistency regularization against soft pseudo-label $r^{w}=\hat{\mathcal{N}}(\hat{\theta}(x^{w}))$ in the low-confidence portion of unlabeled data. In order to validate the effectiveness of learning scheme of TNC in MutexMatch, we use three modified learning schemes for experiments as following:

\begin{enumerate}
  \item [(\romannumeral1)] 
  We use hard pseudo-label  $\hat{q}^{w}=\arg \min (p^{w}) $ to train TNC separately while stopping gradient back-propagation on $\theta$, 
and enforce consistency regularization against hard complementary pseudo-label: 
\begin{equation}\mathcal{L}_{sep}=\frac{1}{\mu B}\sum_{n = 1}^{\mu B}H(\hat{q}^{w}_{n},r^{w}_{n}),\end{equation}
\begin{equation}\mathcal{L}_{ab2}=\frac{1}{\mu B}\sum_{n = 1}^{\mu B}\mathbbm{1}(\max (p^{w}_{n})<  \tau )H(\hat{r}^{w}_{n},r^{s}_{n}),   \end{equation}
where $\hat{r}^{w}=\arg \max (r^{w}) $ and $r^{s}=\mathcal{N}(\theta (x^{s}))$.    
  \item [(\romannumeral2)]
  We use hard complementary pseudo-label $\hat{\gamma}^{w}$, which is generated via randomly selecting the class without the highest confidence from $p^{w} $ (just like the standard complementary label selection) to train TNC separately, while stopping gradient on $\theta$, 
and enforce consistency regularization against soft complementary pseudo-label: 
\begin{equation}\mathcal{L}_{sep}=\frac{1}{\mu B}\sum_{n = 1}^{\mu B}H(\hat{\gamma}^{w}_{n},r^{w}_{n}),\end{equation}
\begin{equation}\mathcal{L}_{ab2}=\frac{1}{\mu B}\sum_{n = 1}^{\mu B}\mathbbm{1}(\max (p^{w}_{n})<  \tau )H({r^{w}_{n}},r^{s}_{n}).  \end{equation}
  \item [(\romannumeral3)]
  We remove the separately training part of TNC. The complementary pseudo-label for TNC is obtained directly by $q^{w}=\texttt{Norm}(\texttt{1}-r^{w})$ where \texttt{1} is a all-one vector and  $\texttt{Norm}(\cdot )$ is operation normalizing $q^{w}$ into interval $[0,1]$. We enforce consistency against soft complementary pseudo-label: 
\begin{equation}\mathcal{L}_{ab2}=\frac{1}{\mu B}\sum_{n = 1}^{\mu B}\mathbbm{1}(\max (p^{w}_{n})<  \tau )H({q^{w}_{n}},r^{s}_{n}).   \end{equation}
\end{enumerate}

The losses minimized by experiments is simply $\mathcal{L}_{sup}+\mathcal{L}_{p}+\mathcal{L}_{sep}+\mathcal{L}_{ab2}$ in (\romannumeral1), (\romannumeral2) and $\mathcal{L}_{sup}+\mathcal{L}_{p}+\mathcal{L}_{ab2}$ in (\romannumeral3). 

\begin{figure}[t] 
    \centering
    
     \subfloat[]{
   \includegraphics[width=4.1cm,height=3.5cm]{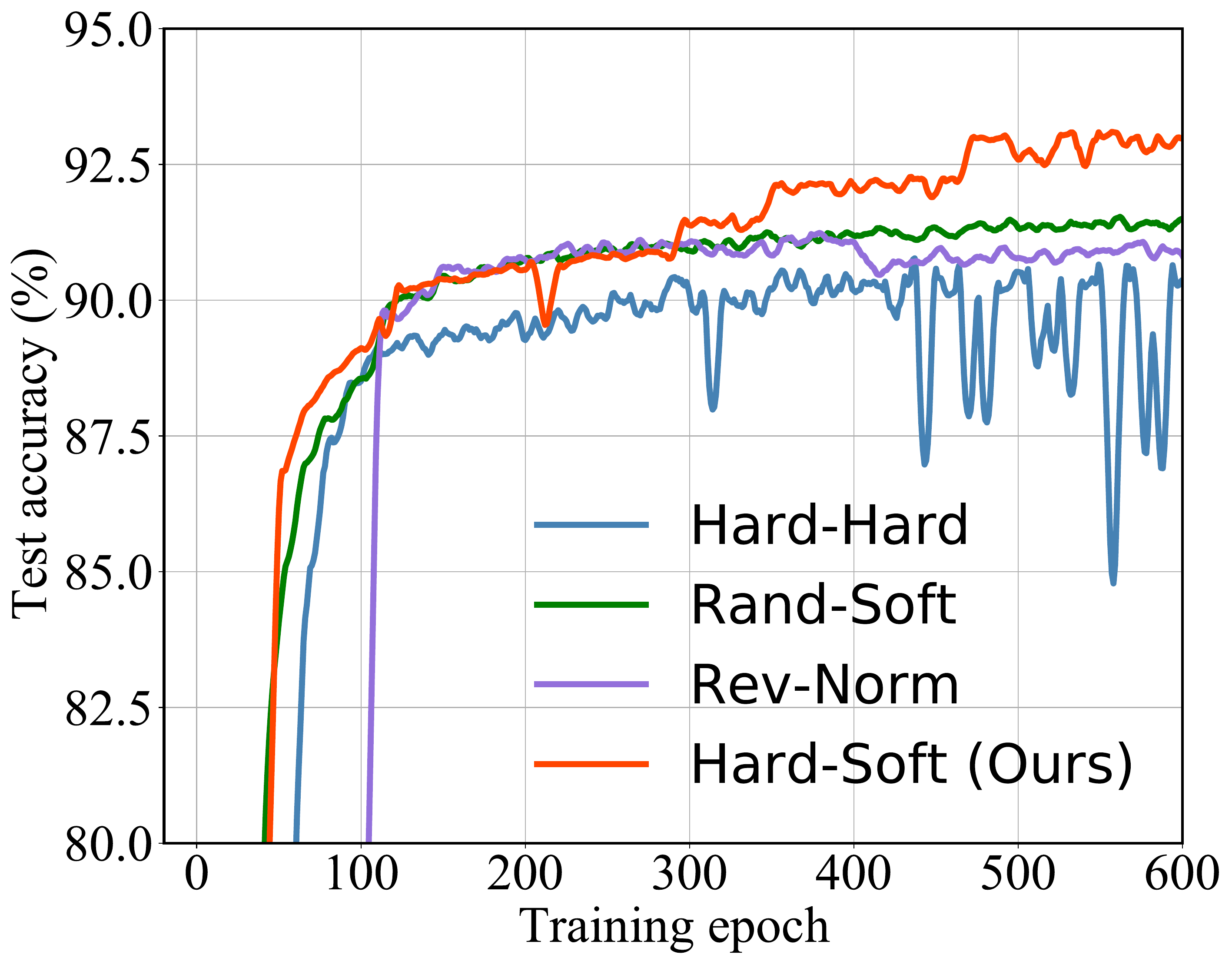}
   }
   \hfil
   \subfloat[]{
   \includegraphics[width=4.1cm,height=3.5cm]{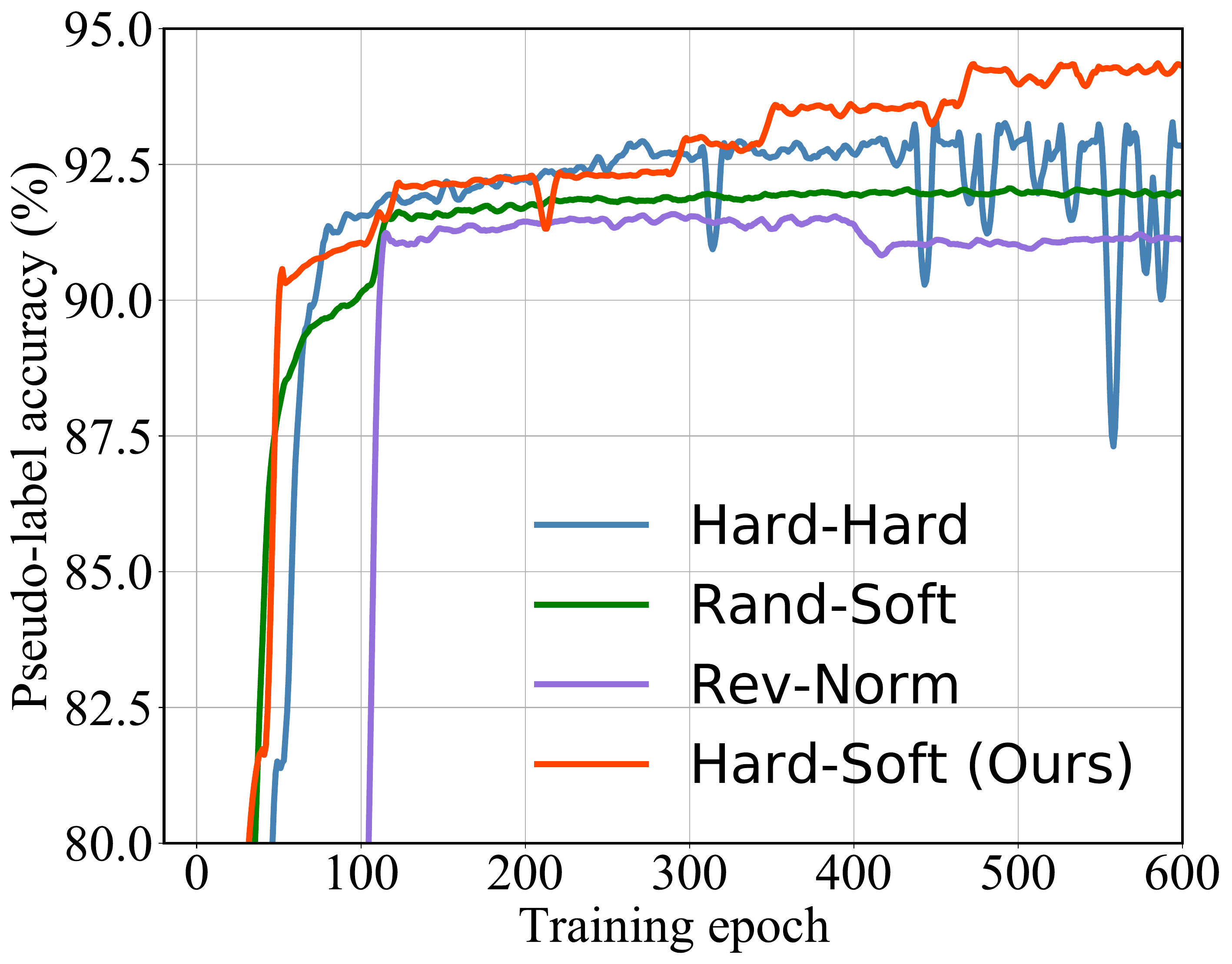}
   }
    \vskip 0in
    \caption{The learning curve of ablation study on CIFAR-10 with 40 labels. The x-axis represents the training epoch while the y-axis represents the test accuracy in (a) and the pseudo-label accuracy in (b).}
    \label{fig3}

  \end{figure}

All models are trained on CIFAR-10 using 4 labels per class, and the results of all experiments are listed in \cref{fig3}.
In the figure, the \textit{Hard-Hard} indicates setting of (\romannumeral1); the \textit{Rand-Soft} indicates setting of (\romannumeral2); and the \textit{Rev-Norm} indicates setting of (\romannumeral3). 
The default MutexMatch achieves an accuracy of 93.22$\pm$2.52\%, outperforming all other settings.
% Furthermore, experiments show that the accuracy and stability of the complementary pseudo-labels which are outputted by TNC of default MutexMatch are dominant. 
MutexMatch uses hard labels for separate training of TNC to ensure a robust classifier that could output more accurate complementary pseudo-labels, 
and uses soft labels to participate in consistency regularization on TNC for informative guidance.

\subsection{Analysis of Consistency Regularization on TNC}
\label{dis_section}
\begin{figure}[t]
    \centering
   \subfloat[]{
   \includegraphics[width=4cm,height=3.5cm]{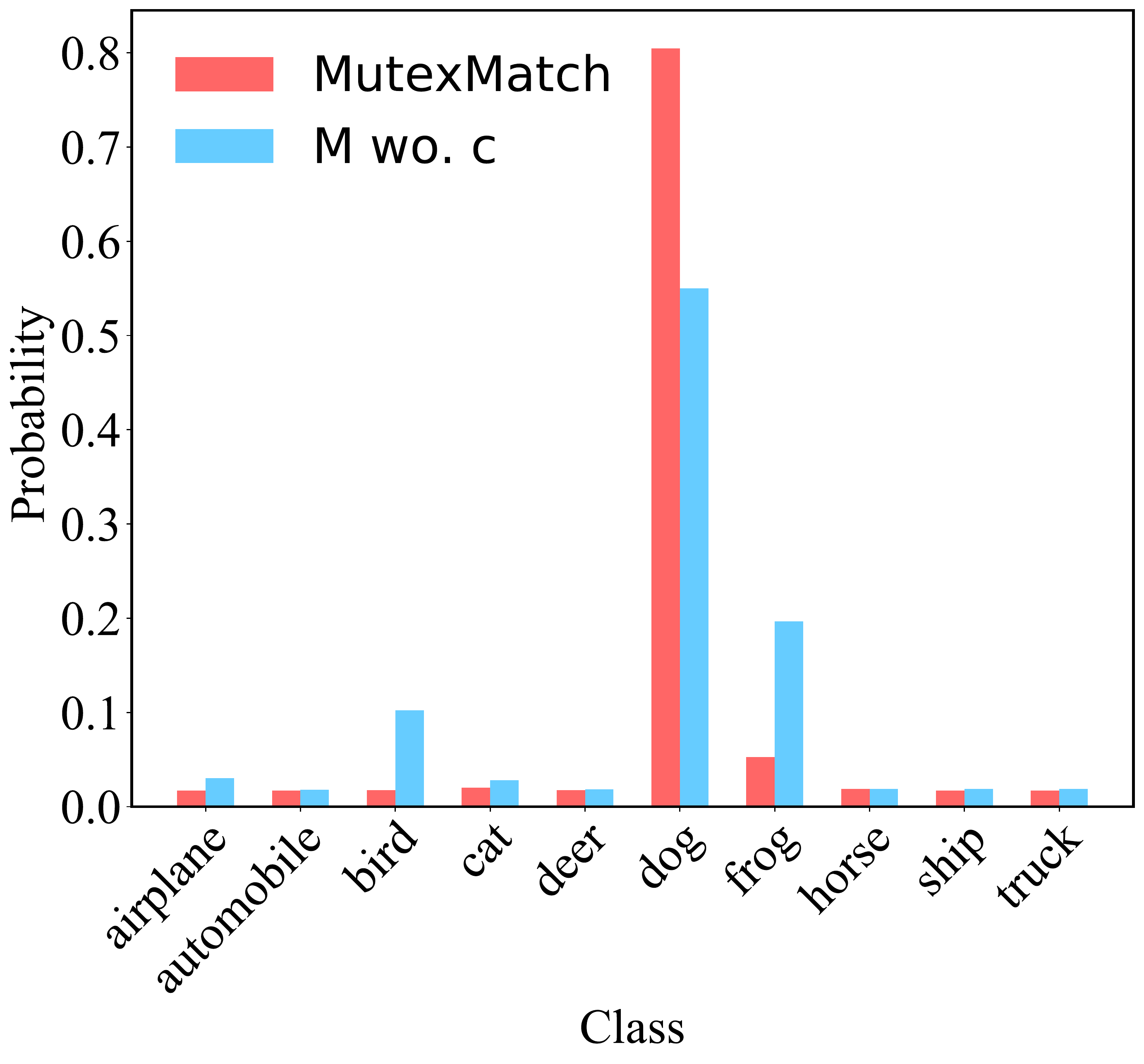}
   \label{fig:re1}
   }
  \hfil
   \subfloat[]{
   \includegraphics[width=4cm,height=3.5cm]{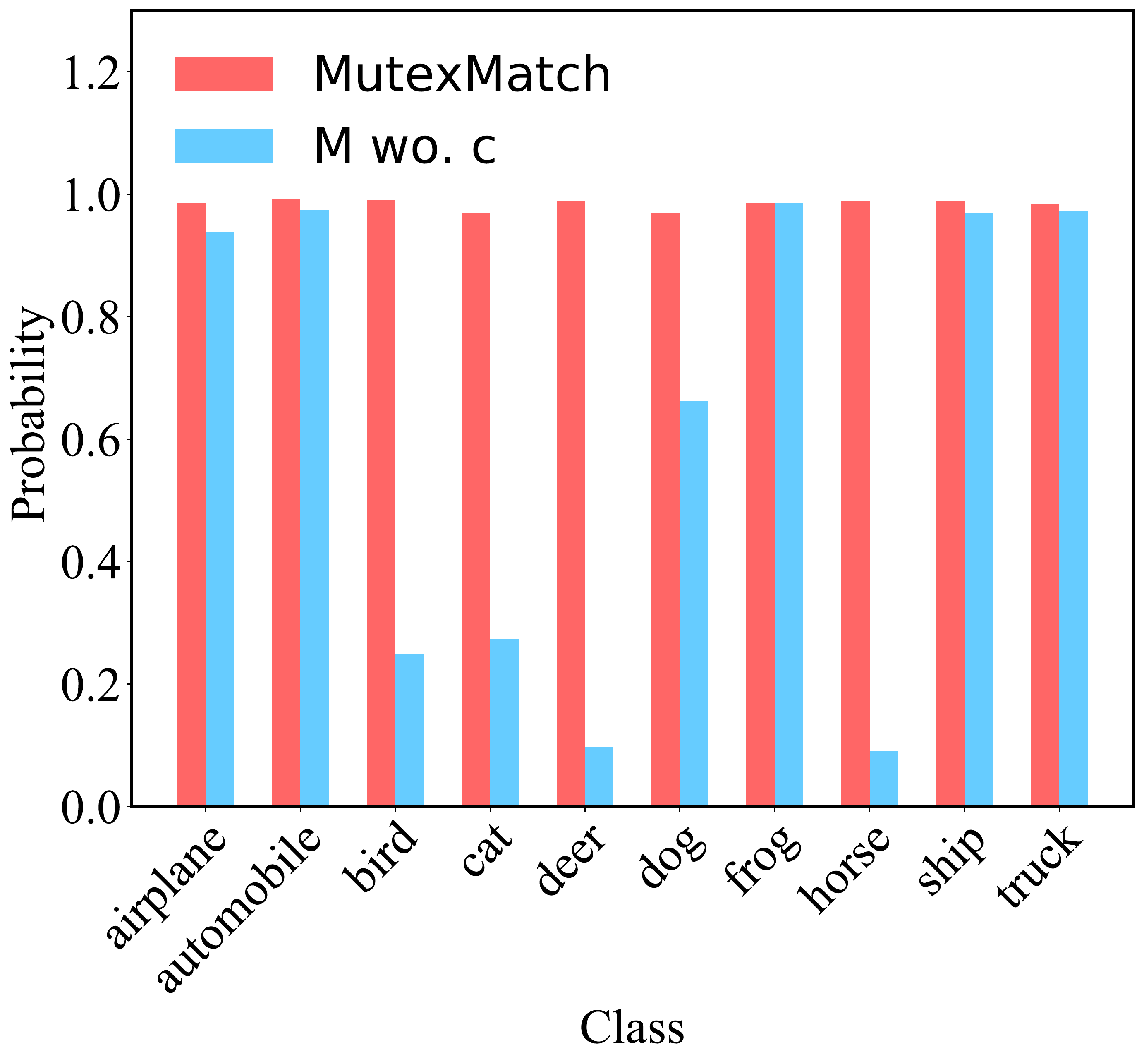}
   
   \label{fig:re2}
   }
%   \vspace{-1em}
  \caption{(a) Average probability of each component in prediction vectors of dog images in CIFAR-10. (b) Average probability of ground-truth component in prediction vectors of all classes in CIFAR-10.}
%   \label{fig2}
  \vskip 0in
\end{figure}

Given two augmented variants derived from the same unlabeled instance, we claim that their predictions of the degree of dissimilarity could share some overlap, which can be achieved by encouraging the class probability distributions (\ie, soft-labels) of their complementary predictions (\ie, the TNC's outputs) to be consistent. 
% We begin with the reason why we enforce consistency regularization on TNC. In brief, considering two versions of augmentation for the same unlabeled data, we claim that the class probability distribution of complementary prediction should be consistent.
Under the help of an independent training process of TNC, the model can be more confident on ``what it is not''. As a result, such prediction consistency on TNC can effectively decrease the False-Negative probability in TPC's predictions,  
% As shown  in \cref{retnc}, 
\eg, given an instance of class ``dog'', the independent training of TNC can generate an accurate complementary prediction with extremely low probability of class ``dog''. Then encouraging a similar prediction on its strongly-augmented variant can help the model to learn more discriminative features, which can in turn affect the TPC's prediction (TNC and TPC share the same backbone) in a way that the False-Negative probabilities (to be predicted as other classes except ``dog'') can be effectively lowered. Consequently, the True-Positive probabilities of TPC's predictions is enlarged.
% encourages classes that may be wrong in the 
% predictions output by the model to have a lower probability, which is illustrates  in \cref{retnc}.
We construct an experiment on CIFAR-10 with 40 labels to verify our findings. We denote MutexMatch without consistency regularization on TNC as \textit{M wo. c} (\ie, MutexMatch degenerates to FixMatch).
We take the predictions from MutexMatch and \textit{M wo. c} as an example for comparison. In order to better display the results, we show the results at 100 epochs.
% In fact, for the correct part of the pseudo-labels, MutexMatch obtains more correct pseudo-labels than \textit{M wo. c} (FixMatch) thanks to the use of consistency regularization on TNC (\ie, MutexMatch's accuracy of pseudo-labels is higher than FixMatch, which is shown  in \cref{ap1}).
As shown  in \cref{fig:re1}, given the unlabeled instances belonging to ``dog'', the MutexMatch's average probability of ``dog'' component in prediction vectors is higher than that of \textit{M wo. c}.
We can also obtain similar findings on other different classes, as shown  in \cref{fig:re2}. 
Such observations demonstrate enforcing prediction consistency on TNC can  help the model lower the False-Negative probabilities, which in turn improve the True-Positive probabilities in the TPC's prediction vectors. 
% More effectiveness analysis of predictions from TNC can be found  in \cref{sec:atnc}.

\section{Conclusion}
\label{sec:con}
In this paper, we propose MutexMatch, a novel SSL algorithm using a mutex-based consistency regularization derived from two distinct classifiers. One is to predict ``what it is'' and the other is to predict ``what it is not''. 
MutexMatch can achieve superior performance on various SSL benchmarks.
% especially under label-scarce conditions. 
Last but not least, we validate that low-confidence samples can also be well utilized in training in a novel way.
We believe this utilization of low-confidence samples can be borrowed to other semi-supervised tasks,
\eg, segmentation and detection.

%\section*{Acknowledgments}
%The work was supportd by National Key Research and Development Program of China (2019YFC0118300), NSFC Major Pro- gram (62192783), CAAI-Huawei MindSpore Project (CAAIXSJLJJ-2021- 042A), China Postdoctoral Science Foundation Project (2021M690609), Jiangsu Natural Science Foundation Project (BK20210224), and Australian Research Council (ARC DP200103223).

% {\appendix[Proof of the Zonklar Equations]
% Use $\backslash${\tt{appendix}} if you have a single appendix:
% Do not use $\backslash${\tt{section}} anymore after $\backslash${\tt{appendix}}, only $\backslash${\tt{section*}}.
% If you have multiple appendixes use $\backslash${\tt{appendices}} then use $\backslash${\tt{section}} to start each appendix.
% You must declare a $\backslash${\tt{section}} before using any $\backslash${\tt{subsection}} or using $\backslash${\tt{label}} ($\backslash${\tt{appendices}} by itself
%  starts a section numbered zero.)}


% \section{References Section}
% You can use a bibliography generated by BibTeX as a .bbl file.
%  BibTeX documentation can be easily obtained at:
%  http://mirror.ctan.org/biblio/bibtex/contrib/doc/
%  The IEEEtran BibTeX style support page is:
%  http://www.michaelshell.org/tex/ieeetran/bibtex/
 

\bibliography{example_paper}
% You can manually copy in the resultant .bbl file and set second argument of $\backslash${\tt{begin}} to the number of references
%  (used to reserve space for the reference number labels box).

\bibliographystyle{IEEEtran}

\vfill

\end{document}